\documentclass[10pt]{article} %
\usepackage[accepted]{rlc}

\usepackage{amssymb}            %
\usepackage{mathtools}          %
\usepackage{mathrsfs}           %
\usepackage{graphicx}           %
\usepackage{subcaption}         %
\usepackage[space]{grffile}     %
\usepackage{url}                %

\usepackage{booktabs}
\usepackage{amsfonts}

\hypersetup{
    colorlinks,
    linkcolor={red!50!black},
    citecolor={blue!50!black},
    urlcolor={blue!80!black}
}

\usepackage{etoolbox}
\makeatletter
\patchcmd{\hyper@makecurrent}{%
    \ifx\Hy@param\Hy@chapterstring
        \let\Hy@param\Hy@chapapp
    \fi
}{%
    \iftoggle{inappendix}{%
        \@checkappendixparam{chapter}%
        \@checkappendixparam{section}%
        \@checkappendixparam{subsection}%
        \@checkappendixparam{subsubsection}%
        \@checkappendixparam{paragraph}%
        \@checkappendixparam{subparagraph}%
    }{}%
}{}{\errmessage{failed to patch}}

\newcommand*{\@checkappendixparam}[1]{%
    \def\@checkappendixparamtmp{#1}%
    \ifx\Hy@param\@checkappendixparamtmp
        \let\Hy@param\Hy@appendixstring
    \fi
}
\makeatletter

\newtoggle{inappendix}
\togglefalse{inappendix}

\apptocmd{\appendix}{\toggletrue{inappendix}}{}{\errmessage{failed to patch}}

\newcommand{\cvec}[1]{\boldsymbol{\mathrm{#1}}}

\newcommand{\KL}[2]{\textrm{KL}\left[ {#1} \parallel {#2} \right]}

\newcommand{\vours}{V-CoRAL}
\newcommand{\pours}{P-CoRAL}
\newcommand{\sameloss}{Same-Loss}

\newcommand\blfootnote[1]{%
  \begingroup
  \renewcommand\thefootnote{}\footnote{#1}%
  \addtocounter{footnote}{-1}%
  \endgroup
}

\title{Combining Reconstruction and Contrastive Methods for Multimodal Representations in RL}

\author{Philipp Becker\\
    philipp.becker@kit.edu \\
    Karlsruhe Institute of Technology\\
    FZI Research Center for Information Technology
    \AND
    Sebastian Mossburger \\
    Karlsruhe Institute of Technology\\
    \AND 
    Fabian Otto \\
    Bosch Center for Artificial Intelligence \\
    University of T\"ubingen 
    \AND
    Gerhard Neumann \\
    Karlsruhe Institute of Technology\\
    FZI Research Center for Information Technology
    }

\begin{document}

\maketitle

\begin{abstract}
Learning self-supervised representations using reconstruction or contrastive losses improves performance and sample complexity of image-based and multimodal reinforcement learning (RL).
Here, different self-supervised loss functions have distinct advantages and limitations depending on the information density of the underlying sensor modality.  
Reconstruction provides strong learning signals but is susceptible to distractions and spurious information.
While contrastive approaches can ignore those, they may fail to capture all relevant details and can lead to representation collapse. 
For multimodal RL, this suggests that different modalities should be treated differently based on the amount of distractions in the signal.
We propose \emph{Contrastive Reconstructive Aggregated representation Learning (CoRAL)}, a unified framework enabling us to choose the most appropriate self-supervised loss for each sensor modality and allowing the representation to better focus on relevant aspects.
We evaluate \emph{CoRAL's} benefits on a wide range of tasks with images containing distractions or occlusions, a new locomotion suite, and a challenging manipulation suite with visually realistic distractions. 
Our results show that learning a multimodal representation by combining contrastive and reconstruction-based losses can significantly improve performance and solve tasks that are out of reach for more naive representation learning approaches and other recent baselines.\blfootnote{Code: \url{https://github.com/pbecker93/CoRAL/}, Project Page: \url{https://pbecker93.github.io/coral_pp/}} 
\end{abstract}

\section{Introduction}
Most representation learning approaches for reinforcement learning (RL)~\citep{hafner2019dream,hafner2020mastering,hafner2023mastering,laskin2020curl,lee2020stochastic,yarats2021improving, zhang2020learning,zhu2023repo,deng2022dreamerpro} focus on images.
Here, the challenge lies in compressing relevant information while not getting distracted by potentially irrelevant aspects.
Yet,  most agents in realistic scenarios can directly observe their internal states using sensors in the actuators, inertial measurement units, and force and torque sensors.
Including this low-dimensional and concise proprioceptive sensing in representation learning can improve representation quality and downstream RL performance.
For such multimodal representations, State Space Models~\citep{murphy2012machine} are a natural choice as they lend themselves to accumulating information across multiple sensors and time. Previous works suggest using either reconstruction~\citep{hafner2019learning, hafner2020mastering} or contrastive methods~\citep{hafner2019dream, ma2020contrastive, nguyen2021tpc, srivastava2021core}, both with their individual strengths and weaknesses. 
While reconstruction provides an informative learning signal, it may fail to learn good representations if observations are noisy or contain distracting elements~\citep{zhang2020learning, ma2020contrastive, deng2022dreamerpro}.
In such cases, contrastive methods can ignore irrelevant parts of the observation and still learn valuable representations.  
However, they are prone to representation collapse and often struggle to learn accurate dynamics~\citep{ma2020contrastive}.
We argue that the different properties of sensors, such as images and proprioception, suggest using different self-supervised loss functions for each modality.

\begin{figure}[t]
    \centering
    \resizebox{0.9\textwidth}{!}{
    \includegraphics[width=\textwidth]{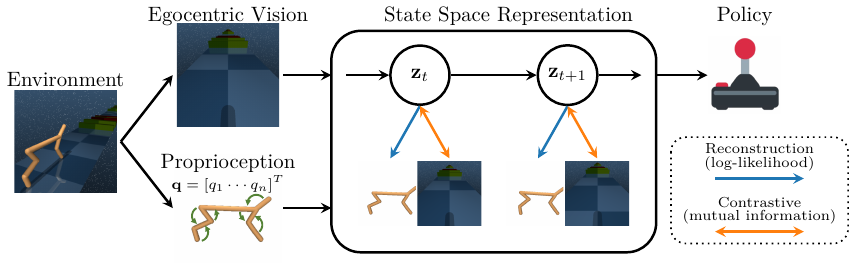}
    }
    \caption{
    \emph{Contrastive Reconstructive Aggregated representation Learning (CoRAL)} learns multimodal state space representations of all available sensors using a combination of reconstruction-based and contrastive objectives.       
    Building on the insight that we can exchange likelihood-based reconstruction with contrastive approaches using mutual information, allows us to choose an appropriate loss function for each modality.  
    Motivated by both a variational and predictive coding viewpoint, \emph{CoRAL} helps model-free and model-based agents to excel in challenging tasks that require information fusion from sensors with different properties such as images and proprioception.     
    }
    \label{fig:fig1}
\end{figure}

We propose \emph{Contrastive Reconstructive Aggregated representation Learning (CoRAL)} to combine contrastive and reconstruction-based approaches.
\emph{CoRAL} builds on state space representations and allows us to select the best-suited loss function for each modality, for example, reconstruction-based loss functions for concise, low-dimensional proprioception and contrastive losses for images with distractions.
Learning such state space representations can be theoretically motivated using a variational inference ~\citep{hafner2019learning, ma2020contrastive} or a predictive coding~\citep{oord2018cpc, nguyen2021tpc, srivastava2021core} viewpoint, which results in two instances of \emph{CoRAL}.
For both paradigms, \emph{CoRAL} relies on the insight that we can replace likelihood-based reconstruction terms with contrastive losses based on mutual information, which allows for a principled combination of the two~\citep{hafner2019dream, ma2020contrastive}. 
\autoref{fig:fig1} provides an overview of the approach.

We integrate \emph{CoRAL} into model-free and model-based RL to systematically assess the effects of learning multimodal representations by selecting appropriate losses.
We evaluate on DeepMind Control (DMC) Suite~\citep{tassa2018deepmind} tasks which we make more difficult by adding \emph{Video Backgrounds}~\citep{zhang2020learning,nguyen2021tpc} and \emph{Occlusions}. 
Furthermore, we use a new \emph{Locomotion} suite where agents must fuse proprioception and egocentric vision to move while navigating obstacles. 
Finally, we consider a novel challenging \emph{Manipulation} suite consisting of static and mobile manipulation tasks with varying object geometries, built on ManiSkill2~\citep{gu2023maniskill2}. 
Here, the agents must combine proprioception and different visual modalities, such as color and depth, to move, navigate, and interact with varying objects in visually realistic environments.
These experiments show that learning multimodal representations using the best-suited loss for each modality improves over other methods combining both modalities, such as representation learning with a single loss and concatenating image representations with proprioception.
\emph{CoRAL} tends to work better than recent baselines on the \emph{Video Background} and \emph{Occlusion} tasks and allows significant performance gains in the challenging \emph{Locomotion} and \emph{Manipulation} tasks. 
Furthermore,  \emph{CoRAL} significantly improves model-based approaches with contrastive image representations, which are known to perform worse than reconstruction-based approaches \citep{hafner2019dream, ma2020contrastive}.
Finally, we show the strengths of both instances of \emph{CoRAL}. \emph{Variational CoRAL} excels in tasks where the main challenge is filtering out irrelevant distractions from images, while \emph{Predictive CoRAL}  performs better in tasks that require propagating information over many timesteps.

To summarize our contributions: \textbf{(i)} We propose \emph{CoRAL}, a general framework for multimodal representation learning for RL which allows using the best-suited self-supervised loss for each modality using the interchangeability of likelihood-based reconstruction and contrastive losses based on mutual information. 
\textbf{(ii)}We instantiate two versions of \emph{CoRAL} using state space representations, namely \emph{Variational-CoRAL} and \emph{Predictive-CoRAL},  which are inspired by variational and contrastive predictive coding viewpoints, respectively. 
\textbf{(iii)} We systematically show their effectiveness on a diverse set of $26$ tasks, across the \emph{Video Backgrounds}, \emph{Occlusions} \emph{Locomotion}, and \emph{Manipulation} suites.

\section{Related Work}

\textbf{Representations for Reinforcement Learning.} 
Many recent approaches use ideas from generative~\citep{wahlstrom2015pixels, watter2015embed, banijamali2018robust, lee2020stochastic, yarats2021improving} and self-supervised representation learning~\citep{zhang2020learning, laskin2020curl, yarats2021reinforcement, stooke2021decoupling, you2022integrating} to improve performance, sample efficiency, and generalization of RL from images.
Those based on \emph{Recurrent State Space Models (RSSMs)}~\citep{hafner2019learning} are particularly relevant for this work. 
When proposing the \emph{RSSM}, \citet{hafner2019learning} used a generative approach.
They formulated their objective as auto-encoding variational inference~\citep{kingma2013auto}, which trains the representation by reconstructing observations.  
Such reconstruction-based approaches have limitations with observations containing noise or many task-irrelevant details. 
As a remedy, \citet{hafner2019dream} proposed a contrastive alternative based on mutual information and the InfoNCE estimator~\citep{poole2019variational}. 
\citet{ma2020contrastive} refined this approach and improved results by modifying the policy learning mechanism. 
Using a different motivation, namely contrastive predictive coding~\citep{oord2018cpc}, \citet{okada2021dreaming, nguyen2021tpc, srivastava2021core, okada2022dreamingv2}  proposed alternative contrastive learning objectives for \emph{RSSMs}. 
In this work, we leverage the variational and predictive coding paradigms and show that \emph{CoRAL} improves performance for both. 
\citet{fu2021tia, wang2022denoised} propose further factorizing the \emph{RSSM}'s latent variable to disentangle task-relevant and task-irrelevant information.
However, unlike contrastive approaches, they explicitly model the task-irrelevant parts instead of ignoring them, which can impede performance if the distracting elements become too complex to model. 
\citet{zhu2023repo} propose a relaxed variational information bottleneck~\citep{alemi2016deep} approach which trains \emph{RSSMs} solely by predicting rewards and enforcing posterior predictability using a KL term. 
Other recent approaches for learning \emph{RSSMs} include using prototypical representations~\citep{deng2022dreamerpro} or masked reconstruction~\citep{seo2022masked}. 

\textbf{Sensor Fusion in Reinforcement Learning.}
Many application-driven approaches to visual RL for robots use proprioception to solve their specific tasks~\citep{finn2016unsupervised, levine2016end, kalashnikov2018scalable, xiao2022masked, Fu_2022_CVPR}.
Yet, they usually do not use explicit representation learning or concatenate image representations and proprioception. 
Several notable exceptions use \emph{RSSMs} with images and proprioception ~\citep{wu2022daydreamer,  becker2022uncertainty, hafner2022director,hafner2023mastering}. Furthermore, \citet{seo2023multi} learn world models using multiple images from different viewpoints.
However, all these approaches focus on purely reconstruction-based representation learning.
\citet{srivastava2021core} use images and proprioceptive information using contrastive predictive coding for both modalities.
Opposed to all of these works, we propose combining contrastive approaches with reconstruction.

\textbf{Multimodal Representation Learning.} 
Representation learning from multiple modalities has widespread applications in general machine learning, where methods such as \emph{CLIP}~\citep{radford2021learning} combine language concepts with the semantic knowledge of images and allow language-based image generation~\citep{ramesh2022hierarchical}.
For robotics, \citet{brohan2022can,mees2022matters,driess2023palm,shridhar2022cliport,shridhar2023perceiver} combine language models with the robot's perception for natural language-guided manipulation tasks using imitation learning.
In contrast, \emph{CoRAL} assumes an online RL setting and focuses on different modalities, namely images and proprioception. 

\section{Combining  Contrastive Approaches and Reconstruction for State Space Represntations}

Given trajectories of observations $\cvec{o}_{1:T} = \lbrace \cvec{o}_t \rbrace_{t=1:T} $ and actions $\cvec{a}_{1:T} = \lbrace \cvec{a}_t \rbrace_{t=1:T}$ we aim to learn a state representation that is well suited for RL.
We assume the observations stem from $K$ different sensors, $\cvec{o}_t = \lbrace\cvec{o}^{(k)}_t\rbrace_{k=1:K}$, where the individual $\cvec{o}_t^{(k)}$ only contain partial information about the state.  
Further, even $\cvec{o}_t$ may not contain all necessary information for optimal acting, i.\,e., the environment is partially observable, and the representation has to accumulate information over time.
Our goal is to learn a concise, low dimensional representation $\phi(\cvec{o}_{1:t}, \cvec{a}_{1:t-1})$ that accumulates all relevant information until time step $t$.
We provide this representation to a policy $\pi (\cvec{a}_t | \phi(\cvec{o}_{1:t}, \cvec{a}_{1:t-1}))$ which aims to maximize the expected return in a given RL problem.
In this setting, the policy's final return and the sample complexity of the entire system determine what constitutes a \emph{good} representation.

State Space Models (SSMs)~\citep{murphy2012machine}  
naturally lend themselves to sensor fusion and information accumulation problems. 
We assume a latent state variable, $\cvec{z}_t$, which evolves according to a Markovian dynamics $p(\cvec{z}_{t+1} | \cvec{z}_t, \cvec{a}_t)$ given an action $\cvec{a}_t$. 
Furthermore, we assume the $K$ observations at each time step are conditionally independent given the latent state, resulting in an observation model
$p(\cvec{o}_t | \cvec{z}_t) = \prod_{k=1}^K p^{(k)}(\cvec{o}^{(k)}_t | \cvec{z}_t)$.
The initial state is distributed according to $p(\cvec{z}_0)$. 
Here, the belief over the latent state, taking into account all previous actions as well as previous and current observations $p(\cvec{z}_t | \cvec{a}_{1:t-1}, \cvec{o}_{1:t})$ can be used as the representation. 
Yet, computing $p(\cvec{z}_t | \cvec{a}_{1:t-1}, \cvec{o}_{1:t})$ analytically is intractable for models of relevant complexity and we use a variational approximation $\phi(\cvec{o}_{1:t}, \cvec{a}_{1:t-1}) \hat{=}q(\cvec{z}_t | \cvec{a}_{1:t-1}, \cvec{o}_{1:t})$. 
This variational approximation also plays an integral part during training and is thus readily available as input for the policy.

We instantiate the generative SSM and the variational distribution using a \emph{Recurrent State Space Model (RSSM)}~\citep{hafner2019learning}, which splits the latent state $\cvec{z}_t$ into a stochastic and a deterministic part.
Following~\cite{hafner2019learning, hafner2019dream}, we assume the stochastic part of the \emph{RSSM}'s latent state to be Gaussian. 
While the original \emph{RSSM} only has a single observation model $p(\cvec{o}_t | \cvec{z}_t)$, we extend it to $K$ models, one for each observation modality.
The variational distribution takes the deterministic part of the state together with the $K$ observations $\cvec{o}_t = \lbrace \cvec{o}_t^{(k)} \rbrace_{k=1:K}$ and factorizes as
$q(\cvec{z}_{1:t} | \cvec{o}_{1:t}, \cvec{a}_{1:t-1}) =  \prod_{t=1}^T q(\cvec{z}_t | \cvec{z}_{t-1}, \cvec{a}_{t-1}, \cvec{o}_t).$
To account for multiple observations instead of one, we first encode each observation individually using a set of $K$ encoders, concatenate their outputs, and provide the result to the \emph{RSSM}.
Finally, we also learn a reward model $p(r_t | \cvec{z}_t)$ to predict the reward from the representation. 
Following the findings of \citet{srivastava2021core} and \citet{tomar2023whatmatter} we also include reward prediction to learn the representations for model-free agents.

\subsection{Learning the State Space Representation}

We propose to combine reconstruction-based and contrastive approaches to train our representations. 
Training \emph{RSSMs} can be based on either a variational viewpoint~\citep{hafner2019dream, ma2020contrastive} or a contrastive predictive coding~\citep{oord2018cpc} viewpoint~\citep{nguyen2021tpc, srivastava2021core}. 
We investigate both approaches, as neither decisively outperforms the other.%

Originally, \cite{hafner2019learning} proposed leveraging a fully generative approach for \emph{RSSMs}. 
Building on the stochastic variational autoencoding Bayes framework~\citep{kingma2013auto, sohn2015learning},
they derive a variational lower-bound objective
\begin{align*}
\mathbb{E}_{p(\cvec{o}_{1:T},  \cvec{a}_{1:T})} \left[\log p(\cvec{o}_{1:T} |  \cvec{a}_{1:T})\right]  \geq \sum_{t=1}^T \mathbb{E}_{\hat{q}(\cdot)} \left[ \log p (\cvec{o}_t | \cvec{z}_t) - \KL{q(\cvec{z}_t | \cvec{z}_{t-1}, \cvec{a}_{t-1}, \cvec{o}_t)}{p(\cvec{z}_t | \cvec{z}_{t-1}, \cvec{a}_{t-1})}\right],
\end{align*}
where $\hat{q}(\cdot) = q(\cvec{z}_{t-1:t} | \cvec{o}_{1:t}, \cvec{a}_{1:t})p(\cvec{o}_{1:t}, \cvec{a}_{1:t})$, i.\,e.,the variational distribution and sub-trajectories from a replay buffer.
After inserting our assumption that each observation factorizes into $K$ independent observations, i.e., $ \log p(\cvec{o}_t | \cvec{z}_t) = \sum_{k=1}^K \log p^{(k)}(\cvec{o}_t^{(k)}| \cvec{z}_t)$,and adding a term for reward prediction, this results in 
\begin{align}
\sum_{t=1}^T \mathbb{E}_{\hat{q}(\cdot)} \left[ \sum_{k=1}^K \log p^{(k)} (\cvec{o}^{(k)}_t | \cvec{z}_t) + \log p(r_t | \cvec{z}_t) -  \KL{q(\cvec{z}_t | \cvec{z}_{t-1}, \cvec{a}_{t-1}, \cvec{o}_t)}{p(\cvec{z}_t | \cvec{z}_{t-1}, \cvec{a}_{t-1})} \right].
\label{eq:recon_bound}
\end{align}
Optimizing this bound using the reparameterization trick~\citep{kingma2013auto, rezende2014stochastic} and stochastic gradient descent simultaneously trains the variational distribution and all parts of the generative model. 
While this approach can be highly effective, reconstructing high-dimensional, noisy observations can also cause issues. 
First, it requires introducing large observation models.
These observation models are unessential for the downstream task and are usually discarded after training.
Second, the reconstruction forces the model to capture all details of the observations, which can lead to highly suboptimal representations if images contain task-irrelevant distractions.

\textbf{Contrastive Variational Learning} (CV) can remedy these problems.   
To introduce contrastive terms, we replace the individual reconstruction terms
in \autoref{eq:recon_bound} with mutual information (MI) terms $I(\cvec{o}_t^{(k)},\cvec{z}_t)$ by adding and subtracting $\log p^{(k)}(\cvec{o}^{(k)})$ \citep{hafner2019dream, ma2020contrastive}
\begin{align}
\mathbb{E}_{\hat{q}(\cdot)}\left[\log p^{(k)}(\cvec{o}_t^{(k)}|\cvec{z}_t)\right] = \mathbb{E}_{\hat{q}(\cdot)}\left[\log 
\frac{p^{(k)}(\cvec{o}_t^{(k)}|\cvec{z}_t)}{ p(\cvec{o}_t^{(k)})} + \log p(\cvec{o}_t^{(k)} ) \right]
= \mathcal{I}(\cvec{o}_t^{(k)},\cvec{z}_t) + c \label{eq:mi_ll}.
\end{align}
Intuitively, the MI measures how informative a given latent state is about the corresponding observations.
Thus, maximizing it leads to similar latent states for similar sequences of observations and actions.
While we cannot analytically compute the MI, we can estimate it using the InfoNCE bound~\citep{oord2018cpc, poole2019variational}.
Doing so eliminates the need for generative reconstruction.
It instead only requires a discriminative approach based on a score function $f_v^{(k)}(\cvec{o}^{(k)}_t, \cvec{z}_t) \mapsto \mathbb{R}_+$.
This score function measures the compatibility of pairs of observations and latent states.
It shares large parts of its parameters with the \emph{RSSM}. 
We refer to \autoref{sup:sec:hp} for details on the exact parameterization. This methodology allows the mixing of reconstruction and mutual information terms for the individual sensors, resulting in a generalization of \autoref{eq:recon_bound}, 
\begin{align}
\sum_{t=1}^T \sum_{k=1}^K \mathcal{L}_v^{(k)}(\cvec{o}^{(k)}_t,  \cvec{z}_t) + \mathbb{E}_{\hat{q}(\cdot)} \left[ \log p(r_t | \cvec{z}_t) - \KL{q(\cvec{z}_t | \cvec{z}_{t-1}, \cvec{a}_{t-1}, \cvec{o}_t)}{p(\cvec{z}_t | \cvec{z}_{t-1}, \cvec{a}_{t-1}}) \right]
\label{eq:cv_bound}.
\end{align}
Here $\mathcal{L}_v^{(k)}$ is either
$\mathbb{E}_{\hat{q}(\cdot)}\left[\log p(\cvec{o}^{(k)}_t | \cvec{z}_t) \right]$ or $\mathcal{I}(\cvec{o}^{(k)}_t, \cvec{z}_t)$.
As we show in \autoref{sec:exp} choosing the terms corresponding to the properties of the corresponding modality can often improve performance. 

\textbf{Contrastive Predictive Coding} (CPC)
\citep{oord2018cpc} provides an alternative to the variational approach. 
The idea is to maximize the MI between the previous latent variable $\cvec{z}_{t-1}$ and the observation $\cvec{o}^{(k)}$, i.\,e., $I(\cvec{o}_t^{(k)}, \cvec{z}_{t-1})$.
While this approach seems similar to contrastive variational learning, we use the previous latent state $\cvec{z}_{t-1}$ instead of the current $\cvec{z}_{t}$ to estimate the MI.
Thus, we explicitly predict one time step ahead to compute the loss. 
As we use the \emph{RSSM's} dynamics model for the prediction, this formalism provides a training signal to the dynamics model. 
However, \citet{levine2019prediction,shu2020predictive, nguyen2021tpc} discuss how this signal alone is insufficient for model-based RL. \cite{srivastava2021core} show that similar ideas also benefit model-free RL and we follow their approach by regularizing the objective using KL-term from \autoref{eq:recon_bound} weighted with a small factor $\beta$.
Additionally, we can turn individual contrastive MI terms into reconstruction terms for suitable sensor modalities by reversing the principle of \autoref{eq:mi_ll}. 
Including reward prediction, this results in the following maximization objective
\begin{align} 
\sum_{t=1}^T \sum_{k=1}^K \mathcal{L}_p^{(k)} (\cvec{o}_{t+1}^{(k)}, \cvec{z}_t) \label{eq:cpc_bound} + \mathbb{E}_{\hat{q}(\cdot)} \left[ \log p(r_t | \cvec{z}_t) - \beta \KL{q(\cvec{z}_t | \cvec{z}_{t-1}, \cvec{a}_{t-1}, \cvec{o}_t)}{p(\cvec{z}_t | \cvec{z}_{t-1}, \cvec{a}_{t-1})} \right],
\end{align}
where $\mathcal{L}_p^{(k)}$ is either the one-step ahead likelihood $\mathbb{E}_{\hat{q}(\cdot)}\left[\log p(\cvec{o}_{t}^{(k)} | \cvec{z}_{t-1}) \right]$ or an InfoNCE estimate of $\mathcal{I}(\cvec{o}_{t}^{(k)}, \cvec{z}_{t-1})$ using a score function $f_p^{(k)}(\cvec{o}^{(k)}_{t}, \cvec{z}_{t-1}) \mapsto \mathbb{R}_+$.
From an implementation viewpoint, the resulting approach differs only slightly from the variational contrastive one. 
For CPC approaches, we use a sample from the \emph{RSSM's} dynamics  $p(\cvec{z}_{t} | \cvec{z}_{t-1}, \cvec{a}_{t-1})$ and for contrastive variational approaches we use a sample from the variational distribution $q(\cvec{z}_t | \cvec{z}_{t-1}, \cvec{a}_{t-1}, \cvec{o}_t)$ as input to the score function or decoder.

\textbf{Estimating Mutual Information with InfoNCE.} We estimate the mutual information (MI) using $b$ mini-batches of sub-sequences of length $l$. 
After computing the latent estimates, we get $N = b\cdot l$ pairs ($\cvec{o}_i$, $\cvec{z}_i$), i.\,e., we use both samples from the elements of the batch as well as all the other time steps within the sequence as negative samples.
Using those, the symmetry of MI, the InfoNCE bound~\citep{poole2019variational}, and either $f=f^{(k)}_v$ or $f=f^{(k)}_p$, we can estimate the MI as
\begin{align*}
\mathcal{I}(\cvec{o}_i, \cvec{z}_i) \geq
   0.5 \left( \sum_{i=1}^N \log \dfrac{f(\cvec{o}_i, \cvec{z}_i)}{\sum_{j=1}^N f(\cvec{o}_i, \cvec{z}_j)} + \log \dfrac{f(\cvec{o}_i, \cvec{z}_i)}{\sum_{j=1}^N f(\cvec{o}_j, \cvec{z}_i)} \right).
\end{align*}

\subsection{Learning to Act Based on the Representation}
Our representations are amenable to both model-free and model-based reinforcement learning. 
For the former, we use Soft Actor-Critic (SAC)~\citep{haarnoja2018sac} on top of the representation by providing the deterministic part of the latent state and the mean of the stochastic part as input to both the actor and the critic. 
For the latter, we use \emph{latent imagination}~\citep{hafner2019dream}, which propagates gradients through the learned dynamics model to optimize the actor. 
In both cases, we alternatingly update the \emph{RSSM}, actor, and critic for several steps before collecting a new sequence in the environment. 
The \emph{RSSM} uses only the representation learning loss and gets neither gradients from the actor nor the critic.

\section{Experiments}
\label{sec:exp}
Building on the previously introduced methodology, we build two versions of \emph{Contrastive Reconstructive Aggregated representation Learning (CoRAL)} differing in the state space representation objective.
\emph{Variational CoRAL} (\emph{V-CoRAL}), using the variational objective (\autoref{eq:cv_bound}) and \emph{Predictive CoRAL} (\emph{P-CoRAL}), using the predictive coding objective (\autoref{eq:cpc_bound}). 
We evaluate the performance of \emph{CoRAL} by using it for downstream online RL and assessing the average expected return or success rate.

To show the benefits of combining contrastive and reconstruction-based objectives, we compare with ablative variants that use the same loss for both modalities (\emph{\sameloss}), the naive approach of concatenating proprioception to image representations (\emph{Concat}) and using only the image (\emph{Img-Only}). 
We consider the contrastive variational (CV) and the contrastive predictive coding (CPC) paradigm for each of these approaches.
For reference, we also include reconstruction-based (Recon.) approaches (\autoref{eq:recon_bound}).
Furthermore, we use SAC~\citep{haarnoja2018sac} on only the proprioception~(\emph{ProprioSAC}), to show that proprioception alone is insufficient to solve the tasks. 
Finally, we consider the model-free \emph{DrQ-v2}~
\citep{yarats2022mastering} and model-based \emph{RePo}~\citep{zhu2023repo} as baselines to demonstrate the competitiveness of our approach. 
We extend both to also use proprioception and refer to the resulting approaches as \emph{DrQ-v2(I+P)} and \emph{RePo(I+P)} respectively. 

\textbf{Evaluation Protocol.}
We run $5$ seeds for each task in each suite and build our analysis on the aggregated results across the entire suite. 
This process results in $35$ runs for each method on \emph{Video Background} and \emph{Occlusions} and $30$ runs for each method in the \emph{Locomotion} and \emph{Manipulation} suites. 
For aggregating the results over a suite, we follow \citet{agarwal2021deep} and provide Interquartile Means (IQMs), which they found to be more meaningful and robust than alternatives such as mean or median in related scenarios. 
Similarly, we follow \citet{agarwal2021deep} and provide 95\% Stratified Bootstrapped Confidence Intervals (CIs) for the entire suite to quantify the statistical uncertainty in results. 
We indicate those with black bars in bar charts or shaded areas in reward curve plots.

\autoref{sup:sec:env} provides details for all tasks. 
\autoref{sup:sec:hp} lists all hyperparameters of our approach and \autoref{sup:sec:baselines} provides further details on the baselines.
\autoref{supp:sec:add_res} shows learning curves for all representation learning paradigms on all tasks, performance profiles, and per-environment results.
Code for running \emph{CoRAL} and ablations on all tasks is available\footnote{\url{https://github.com/pbecker93/CoRAL/}}. 

\subsection{Modified Deep Mind Control Suite Tasks}
\begin{figure}[t]
    \centering
   \includegraphics[width=\textwidth]{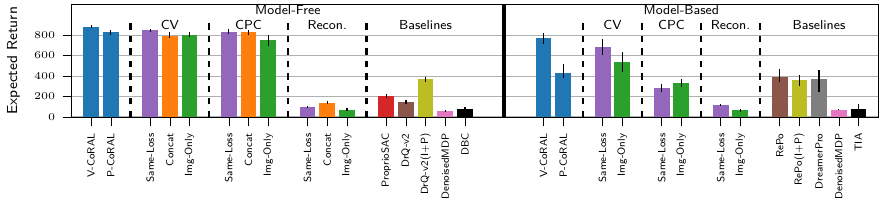}
    \vspace{-0.3cm}
    \caption{Aggregated performance after $10^6$ environment steps on the $7$ tasks from the \emph{Video Background} suite (IQM and $95 \%$ CIs).
    For both model-free and model-based RL, \emph{\vours} performs best among all considered methods, with the model-free performance being better than the model-based one.    
    While some of the model-free ablations are competitive, they perform considerably worse in the model-based case. From the baselines, only \emph{DrQ-v2} with additional proprioception, \emph{RePo} (with and without proprioception), and \emph{DreamerPro} get a final return of over $200$. 
    These results demonstrate how including readily available proprioception with appropriate losses for each modality helps to learn accurate dynamics required by model-based RL and provides a simple alternative to more tailored approaches. 
    }
    \label{fig:main_vid}
\end{figure}
\begin{figure}[t]
\includegraphics[width=\textwidth]{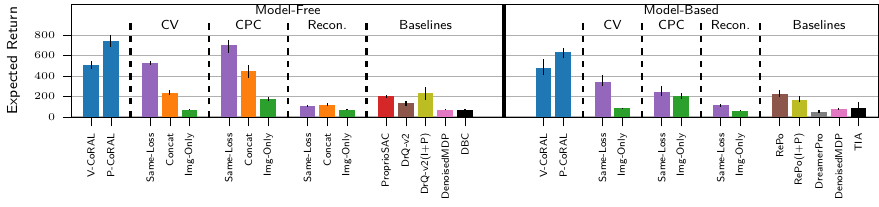}
\vspace{-0.3cm}
\caption{Aggregated performance after $10^6$ environment steps on the $7$ tasks from the \emph{Occlusion} suite (IQM and $95 \%$ CIs).
For both, model-free and model-based RL, \emph{\pours} performs best among all considered methods, with the model-free version again outperforming its model-based counterpart. 
While all approaches handle \emph{Occlusions} worse than \emph{VideoBackgorund}, the performance drop is generally larger for the ablations and baselines. In particular, the \emph{Concat} and model-based \emph{\sameloss} ablations suffer and no approach using only a single modality achieves an expected return of over $200$.
This indicates the importance of learning a multimodal representation using tailored losses over naively integrating proprioception.
}
\vspace{-0.4cm}
\label{fig:main_occ}
\end{figure}

\begin{figure}
\centering
\begin{minipage}[t]{0.55\textwidth}
\includegraphics[width=\textwidth]{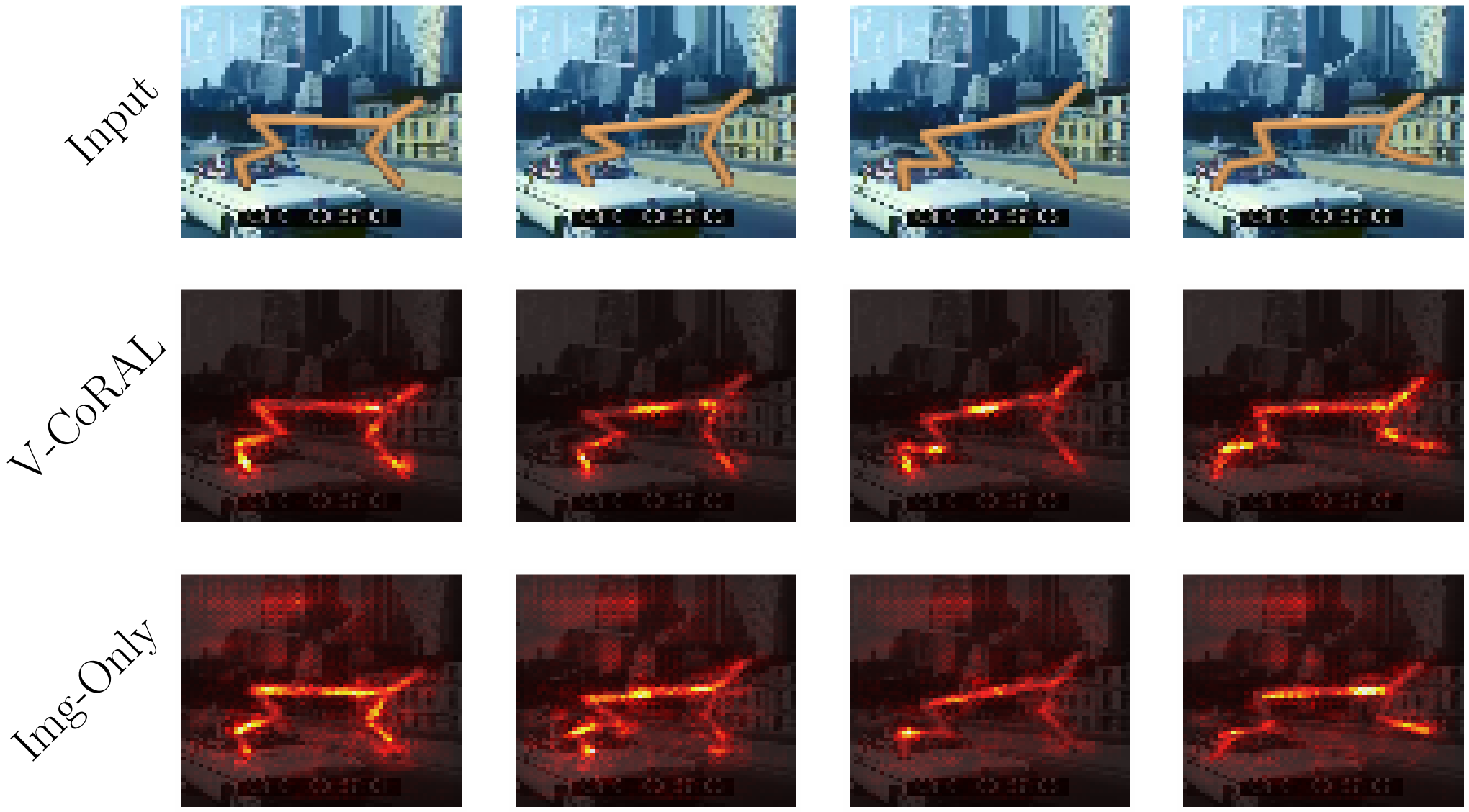}
\end{minipage}%
\begin{minipage}[t]{0.415\textwidth}
\includegraphics[width=\textwidth]{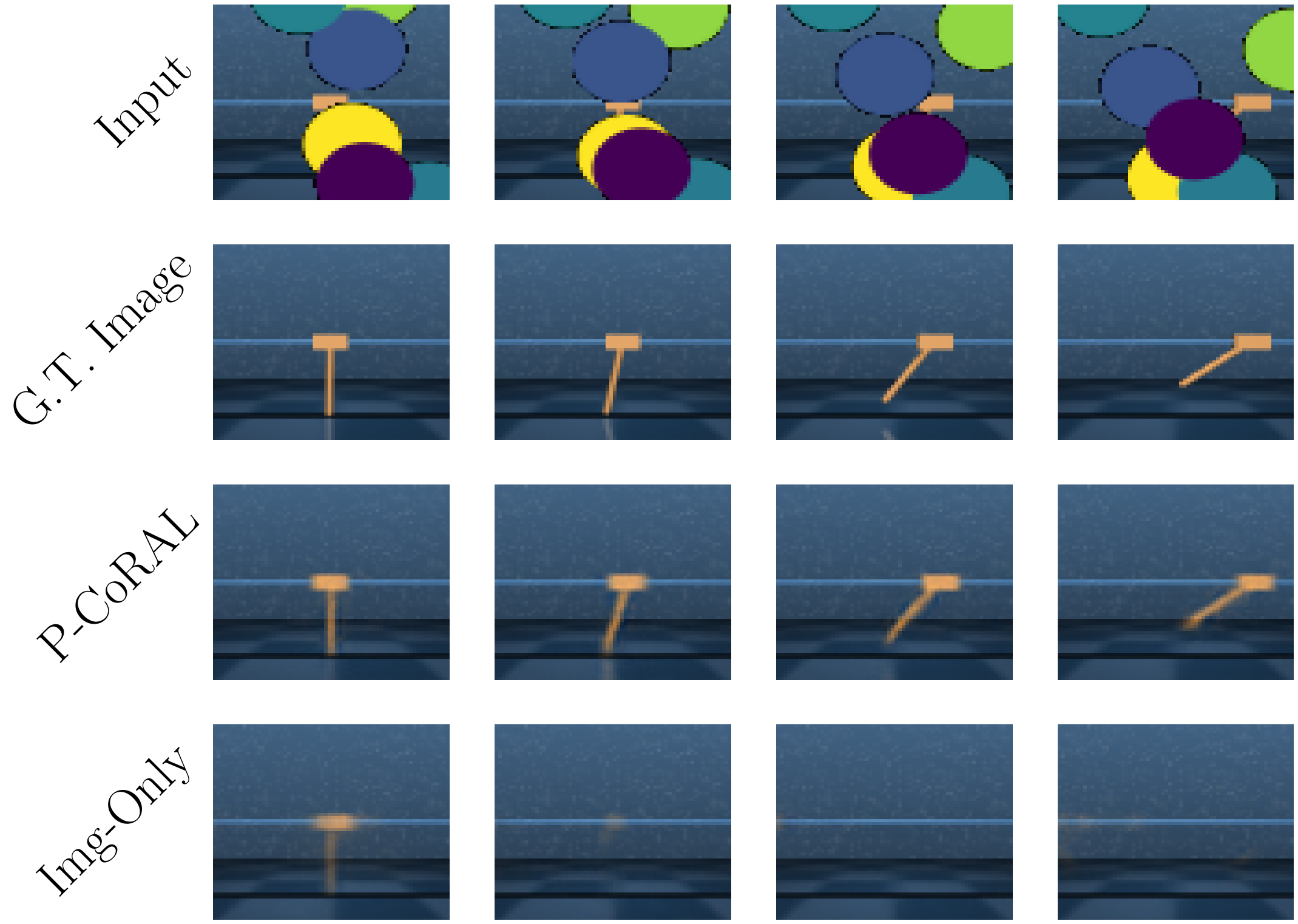}
\end{minipage}
\vspace{-0.3cm}
    \caption{\textbf{Left:} Saliency Maps 
   showing on which pixels the respective representation learning approaches focus in an example from \emph{Video Prediction}. 
   \emph{\vours} focuses better on the task-relevant cheetah, while the corresponding contrastive variational \emph{Img-Only} approach is more distracted by the video background. 
   \textbf{Right:} For this \emph{Occlusion} task, we train a separate decoder to reconstruct the occlusion-free ground truth from the (detached) latent representation.
   For \texttt{Cartpole Swingup} only the cart position is part of the proprioception.
    Still, \emph{\pours} can capture both cart position and pole angle, while the contrastive predictive \emph{Img-Only} approach fails to do so.} 
    \vspace{-0.4cm}
    \label{fig:main_quali_vis}
\end{figure}
We use $7$ tasks from the DeepMind Control Suite (DMC)~\citep{tassa2018deepmind} that cover a wide range of challenges, namely \texttt{Ball-in-Cup Catch}, \texttt{Cartpole Swingup}, \texttt{Cheetah Run}, \texttt{Reacher Easy}, \texttt{Walker Walk}, \texttt{Walker Run}, and \texttt{Quadruped Walk}. 
We split their states into proprioceptive and non-proprioceptive entries, where the proprioception only contains partial information about the state.
The remaining information has to be inferred from images.
For example, in \texttt{Ball-in-Cup Catch} the cup's state is proprioceptive while the ball's state is not. 
\autoref{sup:table:proprioception_dmc} lists the splits for the remaining tasks. 
We create two suites by adding \emph{Video Backgrounds} or \emph{Occlusions} for all seven tasks. For \emph{Video Backgrounds}, we follow~\citep{nguyen2021tpc, deng2022dreamerpro} and render videos from the Kinetics400 dataset \citep{kay2017kinetics} behind the agent. 
For \emph{Occlusions}, we add slowly moving disks in front of the agent.
The upper row of \autoref{fig:main_quali_vis} shows examples. 
For both suites, the challenge is to learn representations that filter out irrelevant visual details while focusing on relevant aspects.
\emph{Occlusions} also tests the approaches' capabilities to maintain a consistent representation across time under partial observability, considerably increasing the task's difficulty. 

For these tasks, we consider the model-free and model-based versions of \emph{\vours}, \emph{\pours}, and all ablative variants. 
Note that the \emph{Concat} ablations are inapplicable in the model-based setting, as the proprioception is not available during \emph{latent imagination}~\citep{hafner2019dream}.
Besides the \emph{DrQ-v2} and \emph{RePo} based baselines, we include several other visual RL approaches tailored for images with distractions to show the competitiveness of \emph{CoRAL}.
Those are the model-based \emph{Task Informed Abstractions (TIA)}~\citep{fu2021tia} and \emph{DreamerPro}~\citep{deng2022dreamerpro},  the model-free \emph{Deep Bisimulation for Control (DBC)}~\citep{zhang2020learning} approach, and 
\emph{DenoisedMDP}~\citep{wang2022denoised}, which has both a model-free and model-based variant.

\autoref{fig:main_vid} and \autoref{fig:main_occ} show the results for the \emph{Video Background} and \emph{Occlusion} tasks respectively. We also include results for the \emph{Standard Images} without any distractors or occlusions for reference and refer to \autoref{supp:sec:add_res} for those results.
On \emph{Natural Videos} \emph{\vours} yields the best results among all approaches.
However, the margin to some of the ablations is small with all of them closing in on the performance of the best approaches on images without background videos (\autoref{fig:mfrl_agg}).
For model-based RL the results show clearer benefits of learning a multimodal representation by appropriately combining multiple losses.
This difference is also much more pronounced for the more difficult \emph{Occlusions}~(\autoref{fig:main_occ}) suite. Here, no image-only approach learns reasonable behavior or manages to outperform \emph{ProprioSAC}, indicating a higher difficulty for representation learning.
Our method \emph{\pours} tends to perform best in this suite, achieving a return of around $750$, and closing in on the best approaches on \emph{Standard Images} which get around $900$.
Furthermore, using readily available proprioception for representation learning in a principled manner provides a simple alternative to the strong baselines listed above and also tends to outperform the naive \emph{Concat} ablation that does not consider proprioception for representation learning but only for RL.

\textbf{Variational vs.\ Predictive Approaches.}
Variational approaches tend to work better than predictive ones on \emph{Video Backgrounds}, where the challenge is to focus on the relevant aspects while ignoring distractions. 
Yet,  the predictive approaches work better on \emph{Occlusions}, where information has to be propagated over time.
As the underlying tasks are identical, this highlights the benefits of considering both paradigms, depending on the perception challenges.

\textbf{Visualization of Learned Representations.} We qualitatively investigate some of the learned representations in \autoref{fig:main_quali_vis}, which illustrates how \emph{CoRAL} helps the representation to focus on relevant aspects and extract all necessary information from an image. 

\textbf{Model Quality and Model-Based Approaches.}
While model-free and model-based agents perform similarly well for approaches that reconstruct images, model-based agents perform worse than their model-free counterparts for contrastive image losses~(\autoref{fig:main_vid},
\autoref{fig:main_occ},
\autoref{fig:mfrl_agg}, \autoref{fig:mbrl_agg}). 
In line with previous findings~\citep{hafner2019dream, ma2020contrastive}, this shows how contrastive approaches struggle to learn suitable long-term dynamics for model-based RL. 
However, this gap is larger for the \emph{\sameloss} and \emph{Img-Only} ablations than for \emph{CoRAL}, which almost closes the gap between model-free and model-based for \emph{\vours}
(\autoref{fig:main_vid}, \autoref{fig:main_occ}).
This result demonstrates how \emph{CoRAL} allows learning more precise long-term dynamics that enable more successful model-based RL. 

\subsection{\emph{Locomotion} Suite}
\begin{figure}[t]

\begin{minipage}{0.45\textwidth}
\vspace{-0.45cm}
\centering
\input{figs/loco_envs/loco_envs}
\end{minipage}%
\begin{minipage}{0.55\textwidth}
\centering
\includegraphics[width=\textwidth]{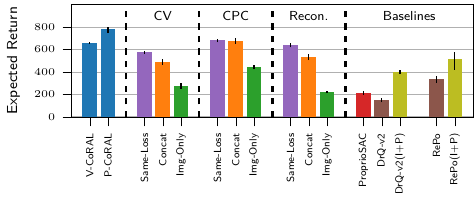}
\end{minipage}
\vspace{-0.4cm}
\caption{\textbf{Left}: Exemplary egocentric (upper row) and external example images (lower row) for the \texttt{Hurdle Cheetah Run}, \texttt{Hurdle Walker Run}, \texttt{Ants Walls}, and \texttt{Quadruped Escape} tasks of the \emph{Locomotion} suite. 
Only the egocentric images are given to the agents, while the external images are solely for visualization of the tasks. 
\textbf{Right:}
Aggregated performance on model-free agents and \emph{RePo} after $10^6$ environment steps on the $6$ tasks of the \emph{Locomotion} suite (IQM and $95 \%$ CIs).
\emph{\pours} significantly outperforms all ablative variants and baselines, highlighting how combining contrastive methods and reconstruction can form effective multimodal representations. It also outperforms purely reconstruction-based approaches, even with no distraction in the images. 
}
\label{fig:main_loco}
\vspace{-0.4cm}
\end{figure}
Building on the DeepMind Control Suite~\cite{Tassa2020dm_control}, we introduce a novel \emph{Locomotion} suite consisting of six tasks: \texttt{Hurdle-Cheetah Run}, \texttt{Hurdle-Walker Walk}, \texttt{Hurdle-Walker Run}, \texttt{Ant-Empty}, \texttt{Ant-Walls} and \texttt{Quadruped Escape}.
All tasks include obstacles that have to be localized through egocentric vision to be avoided. 
As the agents cannot observe themselves from the egocentric perspective, they additionally need proprioception. %
The left side of \autoref{fig:main_loco} provides some examples and \autoref{supp:sec:loco_envs} provides further illustrations and specifications of all tasks. 
These tasks test the representations' ability to combine information from both sources to enable successful navigation and movement. 
For this more challenging suite, we focus on model-free RL for all representations due to the known performance gap for model-based RL with contrastive image losses~\citep{hafner2019dream, ma2020contrastive}, (\autoref{fig:main_vid}, \autoref{fig:main_occ}).
We include the model-based \emph{RePo} for reference.

The results on the right side of \autoref{fig:main_loco} show that \emph{\pours} excels in the \emph{Locomotion} suite and has a significant edge over reconstruction or the pure CPC-based approach while \emph{\vours} outperforms the related variational approaches. While highly relevant to the task, the obstacles appear at random and have random colors for some tasks, which makes reconstruction harder. The contrastive methods' advantage is pronounced in tasks with random colored obstacles~(\autoref{fig:indi:loco}). 

\subsection{\emph{Manipulation} Suite}

\begin{figure}[t]
 \begin{minipage}{0.45\textwidth}
\vspace{-0.45cm}
\centering
\input{figs/replica_envs/replica_envs}
 \end{minipage}%
\begin{minipage}{0.55\textwidth}
\centering
\includegraphics[width=\textwidth]{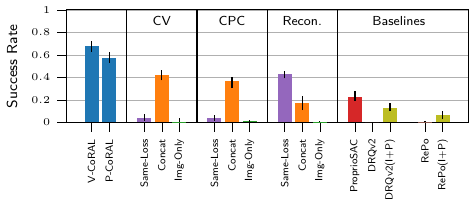}
    \end{minipage}
    \vspace{-0.4cm}
    \caption{
        \textbf{Left}: Exemplary images of the \texttt{LiftCube}, \texttt{TurnFaucet}, \texttt{OpenCabinetDrawer(Depth)}, and \texttt{OpenCabinetDoor(RGBD)} tasks. For the last one, we only show the RGB part of the image and we provide two images per task to showcase the visual diversity and different geometries of the target objects. 
        \textbf{Right:} Aggregated performance on model-free agents and \emph{RePo} after $2\times10^6$ environment steps on the \emph{Manipulation} suite (IQM and $95 \%$ CIs). 
        Overall, \emph{\vours} achieves the best average success rate by a significant margin, followed by \emph{\pours}. 
        While they achieve about $68 \%$ and $58 \%$ success respectively, no ablation gets over $42\%$.
        In particular, both fully contrastive \emph{\sameloss} ablations fail to succeed, which again highlights the importance of choosing an appropriate loss for each modality. 
        While both \emph{RePo} and \emph{DrQ-v2} can utilize the additional proprioception, they are not competitive with \emph{CoRAL} or even SAC trained solely on the proprioception. 
        }
    \label{fig:main_manip}
    \vspace{-0.4cm}
\end{figure}

For the \emph{Manipulation} suite, we design $6$ tasks based on ManiSkill2~\citep{gu2023maniskill2}, i.e., \texttt{LiftCube}, 
\texttt{PushCube},  \texttt{TurnFaucet}, \texttt{OpenCabinetDrawer(RGB)}, \texttt{OpenCanbinetDrawer(Depth)}, and \texttt{OpenCabinetDoor(RGBD)}.
The first three are static manipulation tasks where the target object has to be localized (cube) or identified (faucet) for successful manipulation. 
The latter three are mobile manipulation tasks where the robot navigates to a cabinet and interacts with it using egocentric vision and proprioception. 
They also use different visual modalities, i.e., standard RGB images, depth only, or RGBD. 
For all tasks, we add visually realistic backgrounds using diverse scenes from the ReplicaCAD Dataset~\citep{replica19arxiv} and randomize the ambient lighting. 
The task's complexity stems from the visual realism of the background and the diverse geometry of the target objects, which require that the representations allow identification and precise localization. 
The left side of \autoref{fig:main_manip} provides example images showing the tasks' visual diversity and \autoref{supp:sec:manip_envs} further examples and specifications for all tasks.
We again focus on model-free RL and \emph{RePo}.

The right sight of \autoref{fig:main_manip} shows the results.
The \emph{Manipulation} suite is the hardest set of tasks we consider and here the benefits of \emph{CoRAL} are most obvious. Here, most of the considered baselines fail while only \emph{V-CoRAL} and \emph{P-CoRAL} achieve over $50\%$ success rate, averaged over all tasks, with \emph{V-CoRAL} giving the best result of $68\%$. 
In particular, the corresponding contrastive \emph{same-loss} approaches fail almost completely, which puts additional emphasis on the importance of using appropriate losses for each modality. 
Using different image types for the $3$ mobile manipulation tasks shows how \emph{CoRAL} is beneficial across different visual modalities. 
Using depth images, \texttt{OpenCabinetDrawer(Depth)} effectively removes the lighting variations for this task which allows several approaches to achieve higher performance but has only minor effects on the ranking.
\subsection{Discussion}
Considering all task suites and the full results presented in \autoref{supp:sec:add_res}, we see the benefits of \emph{CoRAL} compared to the ablations and a large selection of model-free and model-based baselines.
Especially for the harder tasks, i.e., \emph{Occlusions}~(\autoref{fig:main_occ}), \emph{Locomotion}~(\autoref{fig:main_loco}), and \emph{Manipulation}~(\autoref{fig:main_manip}), \emph{CoRAL}  can significantly outperform other methods working on the same observations, which shows that different modalities require distinct self-supervised loss functions while simply using the additional proprioception by concatenation or using the same self-supervised loss is often insufficient.

\textbf{Variational vs.\ Predictive Approaches.} 
While either \emph{\vours} or \emph{\pours} generally provides the best results on the considered tasks and both outperform the corresponding ablations, neither consistently outperforms the other across all task suites. 
While this prevents conclusive decisions about whether variational or predictive methods work generally better, we can observe a trend.
The variational approaches appear more suited for tasks requiring the filtering out of visual distractions, such as in \emph{Video Background} and \emph{Manipulation} scenarios, while predictive approaches perform better in tasks needing information to be carried over time, like \emph{Occlusions} and \emph{Locomotion}.

\textbf{Baselines.} 
Contrary to the results presented in the respective original works, 
\emph{DBC}~\citep{zhang2020learning}, \emph{TIA}~\citep{fu2021tia}, \emph{DenoisedMDP}~\citep{wang2022denoised} and \emph{RePo}~\citep{zhu2023repo} underperform on the \emph{Video Backgrounds} task.  
The discrepancy in performance is due to us using a more difficult experimental setup proposed by \cite{deng2022dreamerpro}, which features colored videos of greater diversity. 
We detail the differences and their effects in \autoref{sup:sec:baselines}. 
Furthermore, \emph{RePo} fails in the \emph{Manipulation} suite which seems to contradict results presented by \cite{zhu2023repo} on three static manipulation tasks, similar to those in that suite. 
Again there are subtle differences in the task specification: While \cite{zhu2023repo} only randomize the visual background we randomize both the visual background and the task's initial condition (cube position or faucet model) creating considerably more challenging scenarios.

\textbf{Consistency Across Tasks.}
The additional result visualizations in \autoref{supp:sec:add_res} show that the aggregated performance underlying our analysis is mostly representative of the per-task performance, i.e., if an approach outperforms another when considering the aggregated performance, it generally also does so on a large majority of the individual tasks and runs. 
Furthermore, performance is consistent across the different observation types for the \emph{DMC} tasks, i.e., \emph{Occlusions} are more difficult than \emph{Video Background}, which are more difficult than \emph{Standard Images}~(\autoref{fig:mfrl_agg}, \autoref{fig:mbrl_agg}).

\section{Conclusion}
We consider the problem of Reinforcement Learning (RL) from multiple sensors, in particular images and proprioception. 
We propose \emph{Contrastive reconstructive Aggregated Representation Learning (CoRAL)}, an approach to learning multimodal state space representations for RL by combining contrastive and reconstruction losses.
\emph{CoRAL} builds on the insight that we can replace likelihood-based reconstruction terms with contrastive mutual information terms and vice-versa and is applicable for variational and predictive coding paradigms.
We evaluate on modified versions of the DeepMind Control Suite and novel \emph{Locomotion} and \emph{Manipulation} suites. 
Our results show a consistent benefit of \emph{CoRAL} due to the combination of contrastive approaches for images with reconstruction for low-dimensional, concise signals. These benefits are most pronounced for the hardest tasks we consider, i.e., the \emph{Manipulation} suite, where \emph{CoRAL}, allows us to solve complex tasks with realistic background scenes and varying target object geometries. 

\textbf{Limitations.} 
Depending on the task, either \emph{V-CoRAL} or \emph{P-CoRAL} performs better.
While our evaluation provides some insights about when to use either, further research into understanding their advantages and disadvantages and finding a unified approach that excels in all tasks is required. 
Additionally, even with \emph{CoRAL}, model-free agents outperform their model-based counterparts when using contrastive image losses.
We thus believe that contrastive learning of state space representations can be further improved, especially with regard to learning accurate system dynamics. 

\newpage
\section*{Acknowledgments}
The authors acknowledge support by the state of Baden-Württemberg through bwHPC, as well as the HoreKa supercomputer funded by the Ministry of Science, Research and the Arts Baden-Württemberg and by the German Federal Ministry of Education and Research.
\bibliography{main}
\bibliographystyle{rlc}
\newpage
\appendix

\section{Environments}
\label{sup:sec:env}

\begin{table}[t]
    \centering
        \caption{Splits of the entire system state into proprioceptive and non-proprioceptive parts for the DeepMind Control Suite environments.
    }
    \resizebox{\textwidth}{!}{
    \begin{tabular}{lcc}
    \toprule
    Environment & Proprioceptive & Non-Proprioceptive \\
    \midrule
    Ball In Cup & cup position and velocity &  ball position and velocity \\
    Cartpole & cart position and velocity  & pole angle and velocity   \\
    Cheetah & joint positions and velocities & global pose and velocity \\
    Reacher & reacher position and velocity & distance to target \\
    Quadruped & joint positions and velocities & global pose + velocity, forces \\
    Walker & orientations and velocities of links &  global pose and velocity, height above ground\\
    \bottomrule
    \end{tabular}
    }

    \label{sup:table:proprioception_dmc}
\end{table}

\begin{table}[t]
    \centering
        \caption{Splits of the entire system state into proprioceptive and non-proprioceptive parts for the Locomotion Suite. Some of the agents (Cheetah, Walker, Quadruped) require more proprioceptive information for the locomotion tasks with an egocentric vision than for the standard tasks with images from an external perspective. 
    }
    \begin{tabular}{lcc}
    \toprule
    Environment & Proprioceptive & Non-Proprioceptive \\
    \midrule
    Ant & joint position and velocity & wall positions \\ 
    & global velocities &     global position\\ 
Hurdle Cheetah & joint positions and velocities & hurdle positions and heights \\
     & global velocity &  global position\\ 
Hurdle Walker & orientations and velocities of links  &  hurdle positions and height  \\ 
    & & global position and velocity \\
    Quadruped (Escape)  & joint positions and velocities, & Information about terrain \\ 
     & torso orientation and velocity,  & \\ 
     & imu, forces, and torques at joints &  \\ 
    \bottomrule
    \end{tabular}
    \label{sup:table:proprioception_loco}
\end{table}

\subsection{DeepMind Control Suite Tasks}
\autoref{sup:table:proprioception_dmc} states how we split the states of the original DeepMind Control Suite (DMC)~ \citep{tassa2018deepmind} tasks into proprioceptive and non-proprioceptive parts.
For the model-based agents, we followed common practice \citep{hafner2019dream, fu2021tia, wang2022denoised, deng2022dreamerpro} and use an action repeat of $2$ for all environments. 
We do the same for the model-free agents except for: 
\texttt{Ball In Cup Catch} (4), \texttt{Cartpole Swingup} (8), \texttt{Cheetah Run} (4) and \texttt{Reacher Easy} (4). 
All environments in the locomotion suite also use an action repeat of $2$, this includes \texttt{Hurdle Cheetah Run} which requires more fine-grained control than the normal version to avoid the hurdles.

\paragraph{Natural Background.}
\label{supp:sec:nat_back}
Following \citep{zhang2020learning, fu2021tia, nguyen2021tpc, deng2022dreamerpro, wang2022denoised, zhu2023repo} we render videos from the \texttt{driving car} class of the Kinetics400 dataset \citep{kay2017kinetics} behind the agents to add a natural video background. 
However, previous works implement this idea in two distinct ways. 
\citet{nguyen2021tpc} and \citet{deng2022dreamerpro} use color images as background and pick a random sub-sequence of a random video for each environment rollout. 
They adhere to the train-validation split of the Kinetcs400 dataset, using training videos for representation and policy learning and validation videos during evaluation. 
\citet{zhang2020learning, fu2021tia, wang2022denoised, zhu2023repo}, according to the official implementations, instead work with gray-scale images and sample a single background video for the train set once during initialization of the environment. 
They do not sample a new video during the environment reset, thus all training sequences have the same background video.

We follow the first approach, as we believe it mimics a more realistic scenario of always changing and colored natural background.

\paragraph{Occlusions.} Following \citep{becker2022uncertainty}, we render slow-moving disks over the original observations to occlude parts of the observation.
The speed of the disks makes memory necessary, as they can occlude relevant aspects for multiple consecutive timesteps. 

\subsection{Locomotion Suite}
\label{supp:sec:loco_envs}

\begin{figure}
    \centering
    \input{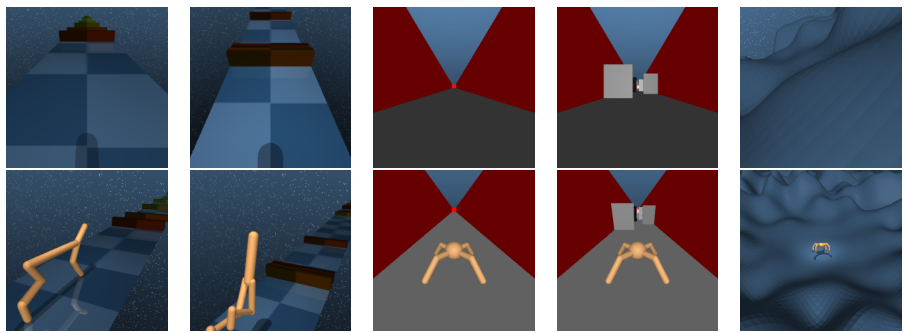}
    \caption{The environments in the Locomotion Suite are (from left to right) \texttt{Hurdle Cheetah Run}, \texttt{Hurdle Walker Walk / Run}, \texttt{Ant Empty}, \texttt{Ant Walls}, and \texttt{Quadruped Escape}. \textbf{Upper Row:} Egocentric vision provided to the agent. \textbf{Lower Row:} External image for visualization.}
    \label{sup:fig:all_loco_vis}
\end{figure}

The $6$ tasks in the locomotion suite are \texttt{Ant Empty}, \texttt{Ant Walls}, \texttt{Hurdle Cheetah Run},
\texttt{Hurdle Walker Walk}, \texttt{Hurdle Walker Run}, and \texttt{Quadruped Escape}.
\autoref{sup:table:proprioception_loco} shows the splits into proprioceptive and non-proprioceptive parts.
\autoref{sup:fig:all_loco_vis} displays all environments in the suite.

Both \textbf{Ant} tasks build on the locomotion functionality introduced into the DeepMind Control suite by \citep{Tassa2020dm_control}. 
For \texttt{Ant Empty}, we only use an empty corridor, which makes this the easiest task in our locomotion suite. 
For \texttt{Ant Walls}, we randomly generate walls inside the corridor, and the agent has to avoid those to achieve its goal, i.e., running through the corridor as fast as possible.

For the \textbf{Hurdle Cheetah} and \textbf{Hurdle Walker} tasks we modified the standard \texttt{Cheetah Run}, \texttt{Walker Walk}, and \texttt{Walker Run} tasks by introducing "hurdles" over which the agent has to step to move forward. 
The hurdles' positions, heights, and colors are reset randomly for each episode, and the agent has to perceive them using egocentric vision. 
For this vision, we added a camera in the head of the Cheetah and Walker. 
Note that the hurdle color is not relevant to avoid them and thus introduces irrelevant information that needs to be captured by reconstruction-based approaches. 

The \textbf{Quadruped Escape} task is readily available in the DeepMind Control Suite. For the egocentric vision, we removed the range-finding sensors from the original observation and added an egocentric camera. 

\subsection{Manipulation Suite}
\label{supp:sec:manip_envs}
\begin{figure}
    \centering
    \input{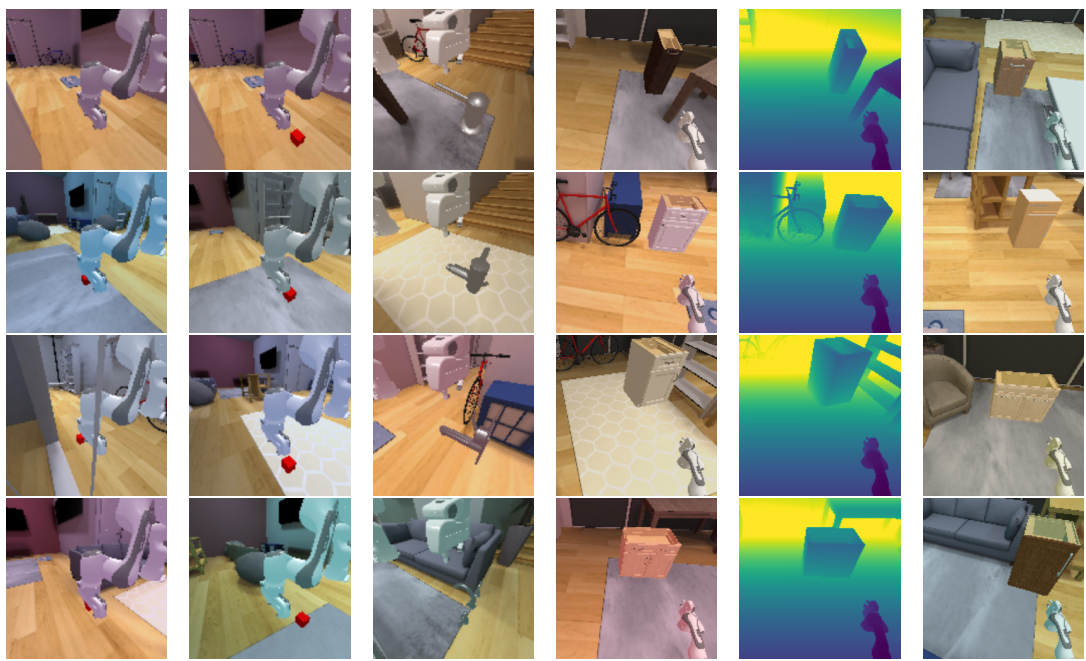}
    \caption{4 Example images for each of the environments in the \emph{Manipulation} Suite, showing the visual and geometric diversity within each task. The tasks are, from left to right, 
    \texttt{LiftCube}, \texttt{PushCube}, \texttt{TurnFaucet}, \texttt{OpenCabinetDrawer(RGB)}, \texttt{OpenCabinetDrawer(Depth)}, \texttt{OpenCabinetDoor(RGBD)}. For the last, we visualize only the RGB part of the image.}
    \label{sup:fig:all_manip_vis}
\end{figure}

The Manipulation Suite builds on Maniskill2~\citep{gu2023maniskill2} and comprises $6$ tasks, i.e., \texttt{LiftCube}, \texttt{PushCube}, \texttt{TurnFaucet}, \texttt{OpenCabinetDrawer(RGB)}, \texttt{OpenCabinetDrawer(Depth)} and \texttt{OpenCabinetDoor(RGBD)}.
The first three involve table-top manipulation and are harder variations of some tasks considered by \cite{zhu2023repo}. 
The latter three are mobile manipulation tasks using different image modalities. 
For all tasks, we use scenes from the Replica Dataset~\cite{replica19arxiv} (specifically: ReplicaCAD\_baked\_lighting\footnote{\url{https://huggingface.co/datasets/ai-habitat/ReplicaCAD_baked_lighting/}}) to place the robot in a visually realistic scene. 
At the beginning of each episode, we randomly pick one of $80$ curated scenes and randomly sample the ambient lighting to place the task in a varying and visually realistic scenery.

We use delta joint position control, no action repeat, and dense normalized rewards for all tasks. 
For the depth images we use the depth camera functionality provided by ManiSkill2 and clip to values between 0 and 4 meters.
Figure \autoref{sup:fig:all_manip_vis} shows example images for all environments. 

\textbf{LiftCube} builds on Maniskill2's LiftCube task and involves picking up a cube and lifting it to a fixed target position. 
The proprioception includes the robot's joint positions, velocities, and end-effector pose, while the cube has to be localized and tracked via an image of an external camera. 
Opposed to \citet{zhu2023repo} we randomize the initial cube position, requiring the agents to first localize the cube based on the representation, which makes the task considerably more difficult. 

\textbf{PushCube} builds on the PushCube task introduced by \cite{zhu2023repo}, but we again randomize the initial cube position. 
Like in LiftCube, the proprioception includes the robot's joint positions, velocities, and end-effector pose, while the cube has to be localized and tracked via an image of an external camera.

\textbf{TurnFaucet} extends Maniskill2's TurnFaucet task and involves opening various faucets by turning the handle. 
The proprioception includes the robot's joint positions, velocities, and end-effector pose, while all information regarding the faucet has to be inferred from an image of an external camera. 
We sample one out of $15$ different faucets at the beginning of each episode. 
As their geometry and opening mechanism vary any representation needs to capture detailed information about the faucet and allow the policy to identify it. 
This makes our task considerably more difficult than that proposed by \citet{zhu2023repo}, who use the same faucet model for all episodes. 

\textbf{OpenCabinetDrawer(RGB)} is based on the mobile manipulation OpenCabinetDrawer task from ManiSkill2, where a mobile robot with a single arm has to navigate towards and then open a drawer one of $25$ cabinets. We disable the rotation of the robot base, which prevents the robot from turning away from the cabinet during initial exploration and significantly speeds up learning for all considered approaches. 
This results in a $10$ dimensional action space, consisting of the $x$ and $y$ velocities of the base, desired changes for the $7$ robot joints, and the gripper. 
Images are egocentric from the top of the robot base and the proprioception includes the entries from the ManiSkill2 "state dict".

\textbf{OpenCabinetDrawer(Depth)} is equivalent to OpenCabientDrawer(RGB) but the agent only receives an egocentric depth image instead of a color image. This effectively removes the variation in lighting from the environment.

For \textbf{OpenCabinetDoor(RGBD)} we build on the Maniskill2 task of the same name, use $25$ different cabinet models, and the same action space as for OpenCabientDrawer(RGB). The sensory observations are also equivalent to the Drawer tasks, but we provide both color and depth information.  
While conceptually similar to the Drawer tasks opening the Door is considerably harder, as it requires coordination with the base not just to reach the handle, but also to pull back on it.

\section{Architecture Details and Training}
\label{sup:sec:hp}
We use the same hyperparameters for all experiments based on the DeepMind Control Suite (DMC), i.e., the standard tasks with the different observation types (\emph{Video Background}, \emph{Occlusions} and also \emph{Standard Images}) as well as, the Locomotion Suite.
For the ManiSkill2-based Manipulation Suite, we use a larger model and a more conservative update scheme for actors and critics. 
We use the ELU activation function unless otherwise mentioned. 

\subsection{\emph{Recurrent State Space Model}}
\label{sup:arch:rssm}
We denote the deterministic part of the \emph{RSSM}'s state by $\cvec{h}_t$ and the stochastic part by $\cvec{s}_t$. 
The base-\emph{RSSM} model without parts specific to the objective consists of:
\begin{itemize}
    \item \textbf{Encoders:} $\psi_{\textrm{obs}}^{(k)}(\cvec{o}_t)$, where $\psi_{\textrm{obs}}$ is the convolutional architecture proposed by \citep{ha2018world} and used by \citep{hafner2019learning, hafner2019dream} for image observations.
    For the low-dimensional proprioception, we used $3 \times 400$ fully connected layers for the DMC tasks and $4 \times 512$ fully connected layers Manipulation Suite.
    \item \textbf{Deterministic Path}: $\cvec{h}_t = g(\cvec{z}_{t-1}, \cvec{a}_{t-1}, \cvec{h}_{t-1}) = \textrm{GRU}(\psi_{\textrm{det}} (\cvec{z}_{t-1}, \cvec{a}_{t-1}), \cvec{h}_{t-1})$ \citep{cho2014gru}, where  $\psi_{\textrm{det}}$ is a $2 \times 400$ fully connected NN and the GRU has a memory size of $200$ for the DMC tasks. For the Manipulation Suite  $\psi_{\textrm{det}}$ has $2 \times 512$ units and the GRU a memory size of $400$
    \item \textbf{Dynamics Model}: $p(\cvec{z}_{t+1} | \cvec{z}_t, \cvec{a}_t) = \psi_{\textrm{dyn}}(\cvec{h}_t)$, where  $\psi_{\textrm{dyn}}$ is a $2 \times 400$ units fully connected NN for the DMC tasks and a $2 \times 512$ units fully connected NN for the Manipulation Suite. The network learns the mean and standard deviation of the distribution. 
    \item \textbf{Variational Distribution} $q(\cvec{z}_t | \cvec{z}_{t-1}, \cvec{a}_{t-1}, \cvec{o}_t) = \psi_{\textrm{var}}\left( \cvec{h}_t, \textrm{Concat}\left(\lbrace{\psi^{(k)}_{\textrm{obs}}(\cvec{o}^{(k)}_t) \rbrace}_{k=1:K}\right) \right)$, where  $\psi_{\textrm{var}}$ is a $2 \times 400$ units fully connected NN for the DMC tasks and a $2\times 512$ units fully connected NN for the Manipulation Suite. 
    Again, the network learns the mean and standard deviation of the distribution. 
    \item \textbf{Reward Predictor} $p(r_t | \cvec{z}_t)$: $2 \times 128$ units fully connected NN for model-free agents. $3 \times 300$ units fully connected NN with ELU activation for model-based agents. The network only learns the mean of the distribution. The standard deviation is fixed at $1$. The model-based agents use a larger reward predictor as they rely on it for learning the policy and the value function. Model-free agents use the reward predictor only for representation learning and work with the ground truth rewards from the replay buffer to learn the critic. 
\end{itemize}

\subsection{Objectives}

\textbf{Image Inputs and Augmentation.} 
Whenever we use a contrastive image loss, we randomly crop a  $64 \times 64$ pixel image from the original image of size $76 \times 76$ pixels during training.  
Cropping is temporally consistent, i.e., the same crop is used for all time steps in a sub-sequence. 
For evaluation, we corp at the center.
For the ablations that reconstruct images, we downsize them directly to $64 \times 64$ pixels.

\textbf{KL.} For the KL terms in \autoref{eq:recon_bound} and \autoref{eq:cv_bound} we follow~\cite{hafner2023mastering} and combine the KL-Balancing technique introduced in~\cite{hafner2020mastering} with the \emph{free-nats regularization} used in~\cite{hafner2019learning, hafner2019dream}. Following~\cite{hafner2020mastering} we use a balancing factor of $0.8$. We give the algorithm $1$ free nat for the DMC Tasks and $3$ for the Manipulation Suite.

\textbf{Contrastive Variational Objective.} 
The score function for the contrastive variational objective is given as
$$f_v^{(k)}(\cvec{o}^{(k)}_t, \cvec{z}_t) = \exp \left( \frac{1}{\lambda} \rho_o \left( \psi_{\textrm{obs}}^{(k)}(\cvec{o}^{(k)}_t)\right)^T\rho_z(\cvec{z}_t) \right),$$
where $\psi_{\textrm{obs}}^{(k)}$ is the \emph{RSSM}'s encoder and $\lambda$ is a learnable inverse temperature parameter. $\rho_o$ and $\rho_z$ are projections that project the embedded observation and latent state to the same dimension, i.e., $50$. $\rho_o$ is only a single linear layer while $\rho_z$ is a $2 \times 256$ fully connected NN. Both use LayerNorm \citep{ba2016layer} at the output.

\textbf{Contrastive Predictive Objective.}
The score function of the contrastive predictive objective looks similar to the one of the contrastive variational objective. The only difference is that the latent state is forwarded in time using the \emph{RSSMs} transition model to account for the predictive nature of the objective, 
$$f_p^{(k)}(\cvec{o}^{(k)}_{t}, \cvec{z}_{t-1}) = \exp \left( \frac{1}{\lambda} \rho_o \left( \psi_{\textrm{obs}}^{(k)}(\cvec{o}^{(k)}_{t})\right)^T\rho_z(\phi_{\textrm{dyn}}(g(\cvec{z}_{t-1}, \cdot)) \right).$$
We use the same projections as in the contrastive variational case.

Following~\cite{srivastava2021core} we scale the KL term using a factor of $\beta=0.001$.

\textbf{Reconstruction Objectives.}
Whenever we reconstruct images we use the up-convolutional architecture proposed by \citep{ha2018world} and used by \citep{hafner2019learning, hafner2019dream}.
For low-dimensional observations, we use $3 \times 400$ units fully connected NN for the DMC tasks and a $4 \times 512$ Units fully connected NN for the Manipulation Suite.
In all cases, only the mean is learned.  We use a fixed variance of $1$ for all image losses and the proprioception for the DMC tasks. For the Manipulation Suite, we set the variance for the proprioception to $0.04$. 

\textbf{Optimizer.} We used Adam~\cite{kingma2015adam} with $ \alpha =3 \times 10^{-4}$, $\beta_1 = 0.99, \beta_2 = 0.9$ and $\varepsilon = 10^{-8}$ for all losses. 
We clip gradients if the norm exceeds $10$.

\subsection{Soft Actor Critic}
\begin{table}[t]
\caption{Hyperparameters used for policy learning with the Soft Actor-Critic.}
\centering
\begin{tabular}{lcc}
\toprule
Hyperparameter & DMC and Locomotion & Manipulation \\
\midrule
Actor Hidden Layers & $ 3 \times 1,024 $ Units & $ 3 \times 1,024 $ Units \\
Actor Activation  & ELU & ELU + LayerNorm\\ 
Critic Hidden Layers & $ 3 \times 1,024 $ Units & $ 3 \times 1,024 $ Units \\
Critic Activation & Tanh & ELU + LayerNorm \\ 
\midrule
Discount & $0.99$ & $0.85$ \\ 
\midrule
Actor Learning Rate & $0.001$  & $0.0003$ \\
Actor Gradient Clip Norm & $10$  &$10$  \\ 
Critic Learning Rate & $0.001$ & $0.0003$  \\
Critic Gradient Clip Norm & $100$ & $100$   \\
\midrule
Target Critic Decay & $0.995$  & $0.995$  \\
Target Critic Update Interval & $1$ & $1$\\
\midrule
$\alpha$ learning rate & $0.001$ & $0.0003$  \\
initial $\alpha$ & $0.1$ & $1.0$\\
target entropy  &- action dim & - action dim \\
\bottomrule
\end{tabular}

\label{sup:table:sac_hps}
\end{table}

\autoref{sup:table:sac_hps} lists the hyperparameters used for model-free RL with SAC~\cite{haarnoja2018sac}.

We collected $5,000$ initial steps at random. 
During training, we update the \emph{RSSM}, critic, and actor in an alternating fashion for $d$ steps before collecting a new sequence by directly sampling from the maximum entropy policy. Here, $d$ is set to be half of the environment steps collected per sequence (after accounting for potential action repeats).
Each step uses $32$ subsequences of length $32$, uniformly sampled from all prior experience.

\subsection{Latent Imagination}

\begin{table}[t]
\caption{Hyperparameters used for policy learning with \emph{Latent Imagination.}}
\centering
\begin{tabular}{lc}
\toprule
Hyperparameter & Value \\
\midrule
Actor Hidden Layers & $ 3 \times 300 $ Units \\
Actor Activation & ELU \\ 
Critic Hidden Layers & $ 3 \times 300 $ Units\\
Critic Activation & ELU \\ 
\midrule
Discount & $0.99$ \\ 
\midrule
Actor Learning Rate & $8 \times 10^{-5}$  \\
Actor Gradient Clip Norm & $100$  \\ 
\midrule
Value Function Learning Rate &  $8 \times 10^{-5}$   \\
Value Gradient Clip Norm & $100$   \\
Slow Value Decay & $0.98$  \\
Slow Value Update Interval & $1$ \\
Slow Value Regularizer & $1$ \\ 
\midrule
Imagination Horizon & $15$  \\
\midrule
Return lambda & $0.95$\\
\bottomrule
\end{tabular}

\label{sup:table:li_hps}
\end{table}

\autoref{sup:table:li_hps} lists the hyperparameters used for model-based RL with latent imagination. They follow to a large extent those used in \cite{hafner2019dream, hafner2020mastering}.

We again collect $5,000$ initial steps at random. 
During training, we update the \emph{RSSM}, value function, and actor in an alternating fashion for $100$ steps before collecting new sequences.
Each step uses $50$ subsequences of length $50$, uniformly sampled from all prior experience.
For collecting new data, we use constant Gaussian exploration noise with $\sigma=0.3$.

\section{Details on Baselines and Ablations.}

\label{sup:sec:baselines}
For \emph{Dreamer-v3}~\citep{hafner2023mastering} we use the raw reward curve data provided with the official implementation\footnote{\url{https://github.com/danijar/dreamerv3/blob/main/scores/data/dmcvision_dreamerv3.json.gz}}.
For \emph{DreamerPro}~\citep{deng2022dreamerpro}\footnote{\url{https://github.com/fdeng18/dreamer-pro}}, \emph{Task Informed Abstractions}~\citep{fu2021tia}\footnote{\url{https://github.com/kyonofx/tia/}}, 
\emph{Deep Bisumlation for Control}~\citep{zhang2020learning}\footnote{\url{https://github.com/facebookresearch/deep_bisim4control/}},
\emph{DenoisedMDP}~\citep{wang2022denoised}\footnote{\url{https://github.com/facebookresearch/denoised_mdp}} and \emph{DrQ-v2}~\citep{yarats2022mastering}\footnote{\url{https://github.com/facebookresearch/drqv2}} we use the official implementations provided by the respective authors. 

\emph{DrQ-(I+P)} builds on the official implementation and uses a separate encoder for the proprioception whose output is concatenated to the image encoder's output and trained using the critics' gradients. 

We implemented \emph{RePo} and \emph{RePo(I+P)} in our framework, reused the Hyperparameters form \cite{zhu2023repo}, and ensured the results of our implementation match the official implementation's\footnote{\url{https://github.com/zchuning/repo}} result on the DMC tasks with standard images. \emph{RePo(I+P)} encodes the proprioception using a separate encoder and both the embedded image and proprioception are given to the \emph{RSSM}.

\textbf{Ablations that are Similar to related Approaches.}
Some of our ablations are very similar to related approaches. 
The model-based \emph{Img-Only} ablation with reconstruction loss, is very similar to \emph{Dreamer-v1}\citep{hafner2019dream}.
It differs from the \emph{Dreamer-v1}~\citep{hafner2019dream} in using the KL-balancing introduced in \citep{hafner2020mastering} and in regularizing the value function towards its own exponential moving average, as introduced in~\citep{hafner2023mastering}. 

However, there are considerable differences between the contrastive version of Dreamer-v1\citep{hafner2019dream} and the contrastive variational \emph{Img-Only} ablation. 
In particular, those regard the exact form of mutual information estimation and the use of image augmentations.

The model-free contrastive predictive \emph{Img-Only} and \emph{\sameloss} ablations are similar to the approach of \cite{srivastava2021core}.
The main difference is that \citet{srivastava2021core} includes the critic's gradients when updating the representation while in our setting no gradients flow from the actor or the critic to the representation. 
Furthermore, we did not include the inverse dynamics objective used by \citet{srivastava2021core} as we did not find it to be helpful. 
Additionally, we adapted some hyperparameters to match those of our other approaches.

\subsection{Hyperparameters of Abltions and Baselines.}

\textbf{Ablations.} All \emph{\sameloss}, \emph{Concat}, and \emph{Img-Only} use the hyperparameters listed in \autoref{sup:sec:hp}. 
They are merely missing certain parts of the model or use a different loss for one or both modalities. 
For the \emph{Concat} baseline, we project the proprioception to the \emph{RSSMs} latent state size (stochastic + deterministic) using a single linear layer before concatenation.  

\emph{ProprioSAC} uses the hyperparameters listed in \autoref{sup:table:sac_hps}, except for the learning rates. 
We reduced those to the SAC default values of $0.0003$ for all environments, as we found the more aggressive updates used for \emph{CoRAL} on \emph{Video Background}, \emph{Occlusions} and \emph{Locomotion} can lead to instabilities when training directly on the proprioception.

\textbf{Baselines.} 
All our baselines were originally evaluated on standard DeepMind Control Suite tasks, modified DeepMind Control Suite tasks, or both. 
They were designed for problems very similar to \emph{Occlusions} and, in particular, \emph{Video Background} and we thus reuse the Hyperparameters originally proposed by the respective authors.
For baselines using an \emph{RSSM}, (\emph{TIA}, \emph{DreamerPro}, \emph{DenoisedMDP}, and \emph{RePo}) these are generally very similar and follow \cite{hafner2019dream, hafner2020mastering}.  

For the \emph{Locomotion} suite all approaches, including \emph{CoRAL} and the ablations, use the same Hyperparameters as they use for \emph{Video Backgrounds} and \emph{Occlusions}.

For the \emph{Manipiluation} suite we increased the model sizes of \emph{RePo} following those of \emph{CoRAL}. 
For both the \emph{DrQ-v2}-based and the \emph{RePo}-based baselines we tried a discount factor of $0.85$ and $0.99$ to ensure the performance differences to \emph{CoRAL} is not an artifact of the low discount of $0.85$. 
However, the lower discount worked better for all methods.

\subsection{On the Performance of Some Baselines in our Setting.}
As described in \autoref{supp:sec:nat_back}, there are distinct ways how to select and use the Kinetics400 videos in the existing literature. 
\citet{nguyen2021tpc}, who first introduced the more challenging setting we use, already found \emph{DBC}~\citep{zhang2020learning} to struggle in this setting and our results align with those findings. 

\emph{TIA}~\citep{fu2021tia} and \emph{DenoisedMDP}~\citep{wang2022denoised} factorize the latent variable into $2$ distinct parts and formulate loss functions that force one part to focus on task-relevant aspects and the other on task-irrelevant aspects.
However, the part responsible for the task-irrelevant aspects still has to model those explicitly.
In the more complicated setting with randomly sampled, colored background videos, the \emph{TIA} and \emph{DenoisedMDP} world models underfit and thus fail to learn a good representation or policy.
Contrastive approaches, such as our approach and \emph{DreamerPro}~ \citep{deng2022dreamerpro}, do not struggle with this issue, as they do not have to model task-irrelevant aspects but can learn to ignore them.

\emph{RePo}~\citep{zhu2023repo} was also evaluated on the simpler setting and \citet{zhu2023repo} report an improved performance over \emph{TIA} and \emph{DenoisedMDP}. 
In the more challenging setting, this improvement persists and \emph{RePo} performs similarly to \emph{DreamerPro} (\autoref{fig:main_vid}).   

Furthermore, \citet{zhu2023repo} presents results on ManiSkill2 environments similar to \texttt{LiftCube}, \texttt{PushCube}, and \texttt{TurnFaucet} of our \emph{Manipulation} Suite. 
However, as detailed in \autoref{supp:sec:manip_envs} our Manipulation Suite tasks randomize initial conditions (i.e., cube position or faucet model) which results in significantly more challenging tasks, in which \emph{RePo} seems to struggle. 

\section{Complete Results}
\label{supp:sec:add_res}
\begin{figure}
    \centering
    \begin{minipage}{\textwidth}
    \centering
    Contrastive Variational Representations\\
    \includegraphics{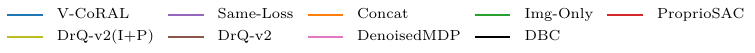}
    \end{minipage}
    \begin{minipage}{\textwidth}
    \centering
    \includegraphics{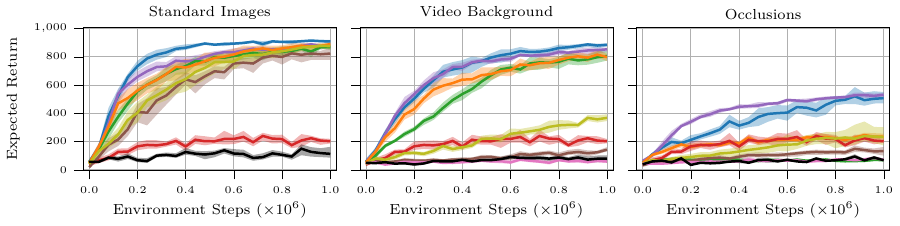}
    \end{minipage}

    \vspace{0.5cm}
    
    \begin{minipage}{\textwidth}    
    \centering
    Contrastive Predictive Coding Representations \\
    \includegraphics{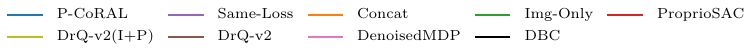}
    \end{minipage}
    \begin{minipage}{\textwidth}
    \centering
   \includegraphics{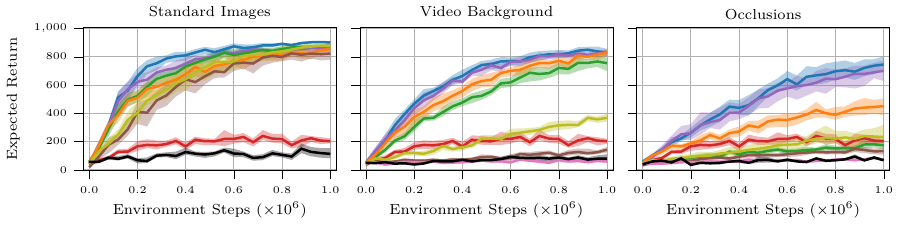}
    \end{minipage}

    \vspace{0.5cm}
    
    \begin{minipage}{\textwidth}
    \centering
    Reconstruction-Based Representations\\
   \includegraphics{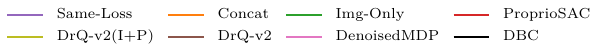}
    \end{minipage}    
    \begin{minipage}{\textwidth}
    \centering
   \includegraphics{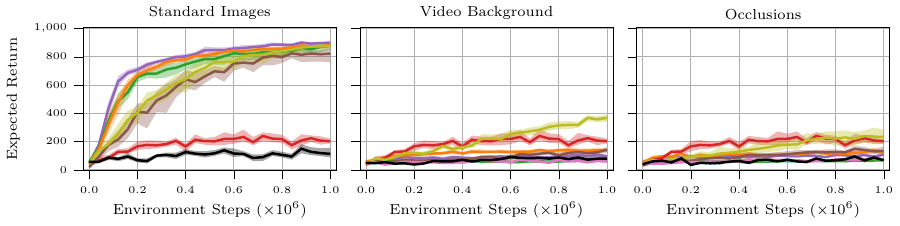}
    \end{minipage}

    \caption{Aggregated results for all \textbf{model-free} agents on the DeepMind Control Suite environments with \emph{Standard Images}, \emph{Video Background}, and \emph{Occlusions}.
     As expected, reconstruction-based approaches do not work on \emph{Video Background} and \emph{Occlusions}.
    Out of all considered approaches \emph{\vours} achieves the highest performance on \emph{Video Background} and \emph{\pours} achieves the highest performance on \emph{Occlusions}.
    }
    \label{fig:mfrl_agg}
\end{figure}

\begin{figure}
    \centering
       
    \begin{minipage}{\textwidth}
    \centering
    Contrastive Variational Representations\\
   \includegraphics{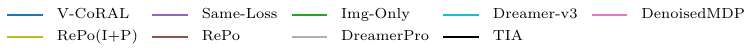}
    \end{minipage}    
    \begin{minipage}{\textwidth}
    \centering
   \includegraphics{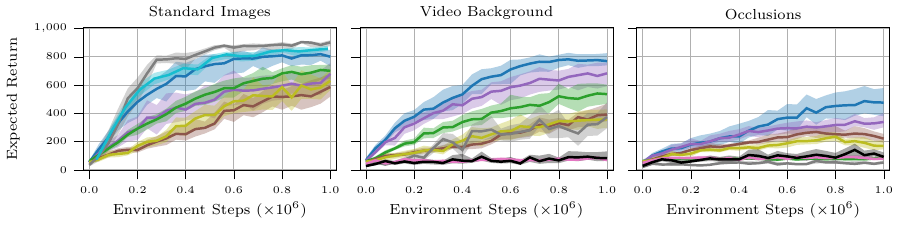}
    \end{minipage}    

    \vspace{0.5cm}
    
    \begin{minipage}{\textwidth}
    \centering
    Contrastive Predictive Coding Representations \\
   \includegraphics{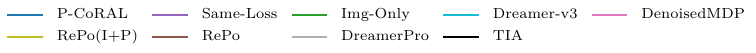}
    \end{minipage}    
    \begin{minipage}{\textwidth}
    \centering
   \includegraphics{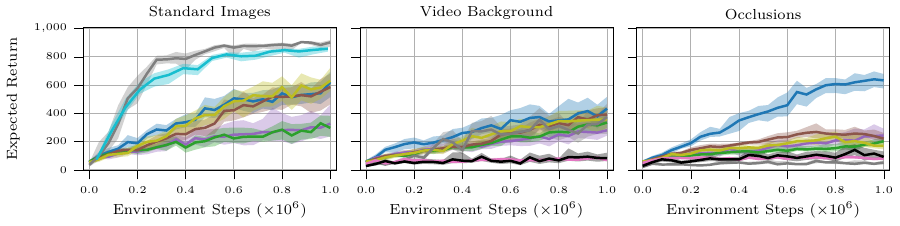}
    \end{minipage}    

    \vspace{0.5cm}
    
    \begin{minipage}{\textwidth}
    \centering
    Reconstruction-Based Representations\\
   \includegraphics{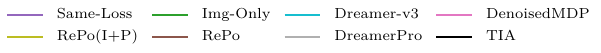}
    \end{minipage}    
    \begin{minipage}{\textwidth}
    \centering
   \includegraphics{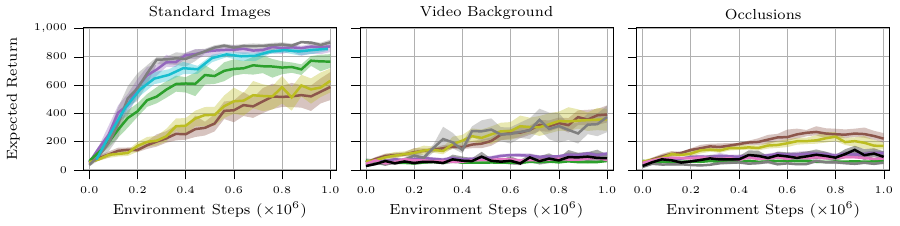}
    \end{minipage}    

    \caption{Aggregated results for all \textbf{model-based} agents on the DeepMind Control Suite environments with \emph{Standard Images}, \emph{Video Background}, and \emph{Occlusions}. 
    Compared to their model-free counterparts~(\autoref{fig:mfrl_agg}), model-based agents perform worse, except if a reconstruction-based representation is used.  
    Yet, the performance gap is larger for image-only and fully contrastive approaches. 
    Especially \emph{\vours} still achieves high performance on \emph{Video Background}, almost matching the performance of \emph{Dreamer-v3} on \emph{Standard Images}.
    This further highlights the benefits of using \emph{CoRAL}, which can significantly improve over tailored approaches such as \emph{DreamerPro} or \emph{RePo}.}
    \label{fig:mbrl_agg}
\end{figure}

\begin{figure}
    \centering
    \begin{minipage}{\textwidth}
    \centering
    Contrastive Variational Representations\\
   \includegraphics{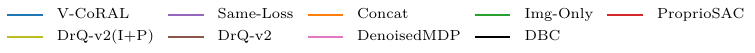}
    \end{minipage}
    \begin{minipage}{\textwidth}
    \centering
   \includegraphics{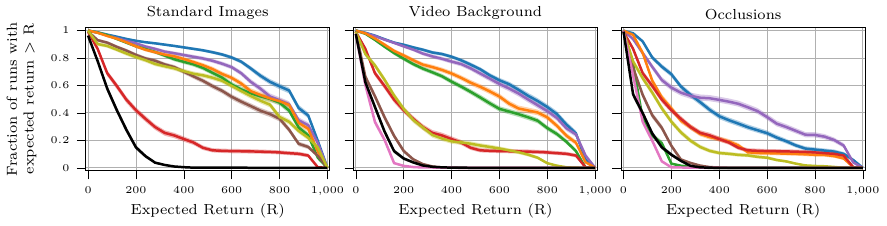}
    \end{minipage}
    
    \vspace{0.5cm}
    
    \begin{minipage}{\textwidth}
    \centering
    Contrastive Predictive Coding Representations \\
   \includegraphics{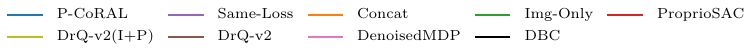}
    \end{minipage}
    \begin{minipage}{\textwidth}
    \centering
   \includegraphics{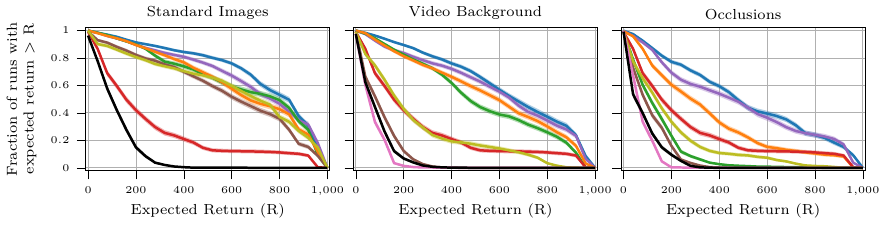}
    \end{minipage}
    
    \vspace{0.5cm}
    
    \begin{minipage}{\textwidth}
    \centering
    Reconstruction-Based Representations\\
   \includegraphics{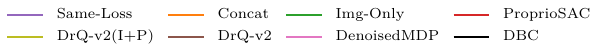}
    \end{minipage}
    \begin{minipage}{\textwidth}
    \centering
   \includegraphics{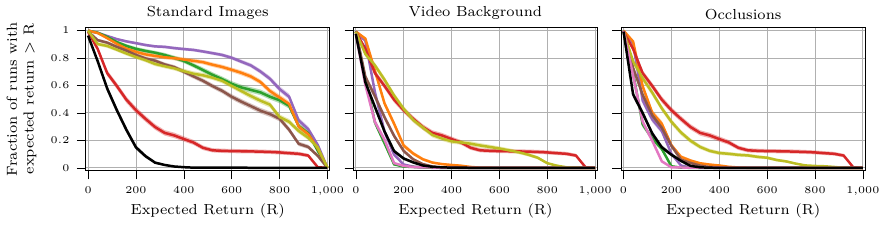}
    \end{minipage}
   \caption{Performance profiles for all \textbf{model-free} agents on the DeepMind Control Suite tasks with \emph{Standard Images}, \emph{Video Background}, and \emph{Occlusions}. They show that performance is largely consistent across the tasks. 
   The sole exception is \emph{\vours} and the contrastive variational approach with the same loss for both modalities on \emph{Occlusions}. Here, the former fails for \texttt{Ball-in-Cup Catch} and \texttt{Cartpole Swingup}, while the latter underperforms for \texttt{Cheetah Run} (\autoref{fig:indi:mfrl_occ}).} 
    \label{fig:mfrl_pp}
\end{figure}

\begin{figure}
    \centering
    \begin{minipage}{\textwidth}
    \centering
    Contrastive Variational Representations\\
   \includegraphics{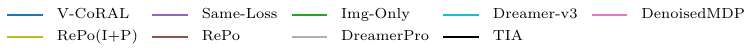}
    \end{minipage}
    \begin{minipage}{\textwidth}
    \centering
   \includegraphics{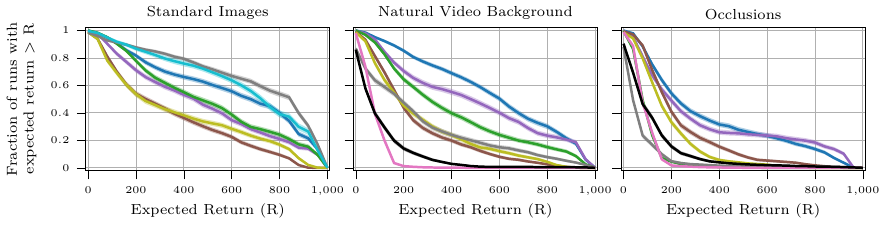}
    \end{minipage}
    
    \vspace{0.5cm}
    
    \begin{minipage}{\textwidth}
    \centering
    Contrastive Predictive Coding Representations \\
   \includegraphics{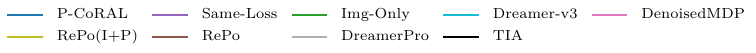}
    \end{minipage}
    \begin{minipage}{\textwidth}
    \centering
   \includegraphics{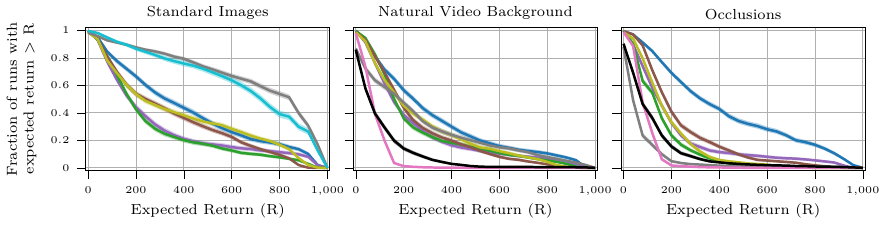}
    \end{minipage}
    
    \vspace{0.5cm}
    
    \begin{minipage}{\textwidth}
    \centering
    Reconstruction-Based Representations\\
   \includegraphics{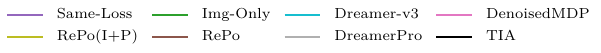}
    \end{minipage}
    \begin{minipage}{\textwidth}
    \centering
   \includegraphics{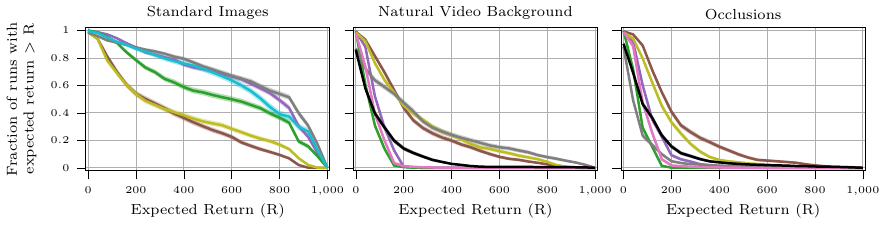}
    \end{minipage}
   \caption{Performance profiles for all \textbf{model-based} agents on the DeepMind Control Suite environments with \emph{Standard Images}, \emph{Video Background}, and \emph{Occlusions}. They indicate that performance is largely consistent across the environments.} 
    \label{fig:mbrl_pp}
\end{figure}

\begin{figure}
    \centering
    \begin{minipage}{\textwidth}
    \centering
    Contrastive Variational Representations\\
   \includegraphics{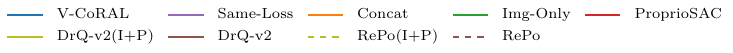}
    \end{minipage}
    \begin{minipage}{\textwidth}
    \centering
   \includegraphics{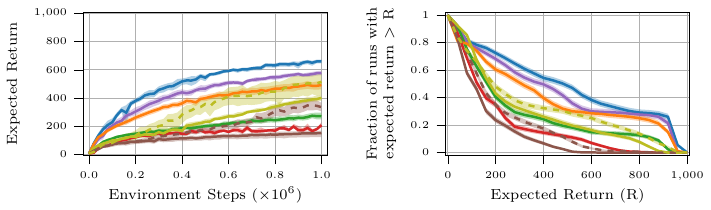}
    \end{minipage}
    
    \vspace{0.5cm}
    
    \begin{minipage}{\textwidth}
    \centering
    Contrastive Predictive Coding Representations \\
   \includegraphics{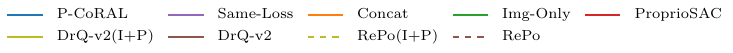}
    \end{minipage}
    \begin{minipage}{\textwidth}
    \centering
   \includegraphics{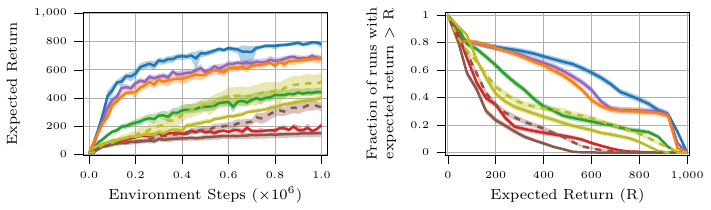}
    \end{minipage}
    
    \vspace{0.5cm}
    
    \begin{minipage}{\textwidth}
    \centering
    Reconstruction-Based Representations\\
   \includegraphics{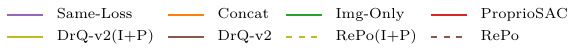}    
    \end{minipage}
    \begin{minipage}{\textwidth}
    \centering
   \includegraphics{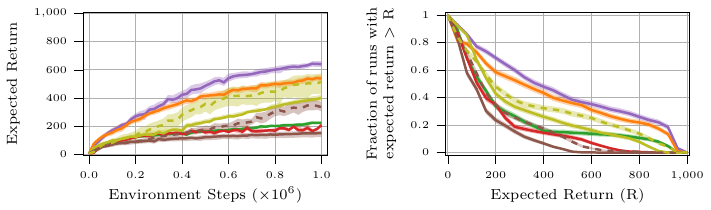}    
    \end{minipage}
    \caption{Aggregated results and performance profiles for the \emph{Locomotion} suite. 
    Both \emph{\vours} and \emph{\pours} outperform reconstruction and \emph{\pours} gives the best results of all approaches by a significant margin 
    \autoref{fig:indi:loco} shows that the performance difference is larger in environments with randomly colored obstacles (\texttt{Hurdle Cheetah Run}, \texttt{Hurdle Walker Walk}, \texttt{Hurdle Walker Run}. 
   The color is not relevant to avoid the obstacles but seems to hinder reconstruction.
 }
    \label{fig:mfrl_loco}
\end{figure}

\begin{figure}
    \centering
    \begin{minipage}{\textwidth}
    \centering
    Contrastive Variational Representations\\
   \includegraphics{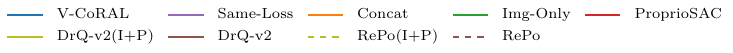}    
    \end{minipage}
    \begin{minipage}{\textwidth}
    \centering
   \includegraphics{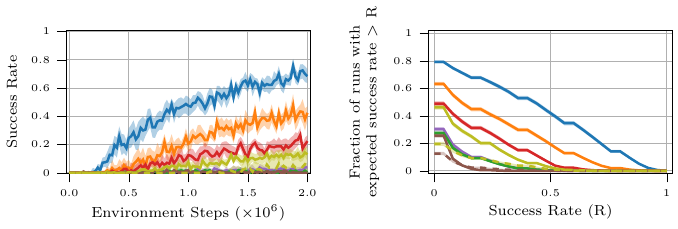}    
    \end{minipage}
    
    \vspace{0.5cm}
    
    \begin{minipage}{\textwidth}
    \centering
    Contrastive Predictive Coding Representations \\
   \includegraphics{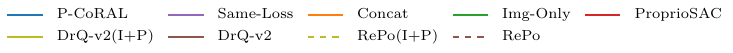}    
    \end{minipage}
    \begin{minipage}{\textwidth}
    \centering
   \includegraphics{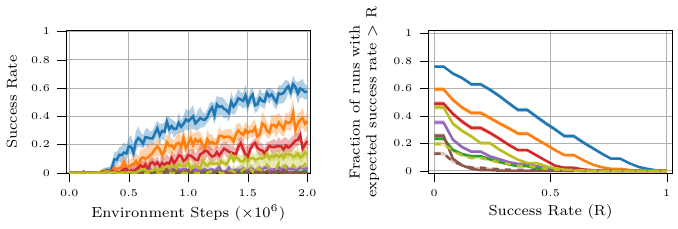}    
    \end{minipage}
        
    \vspace{0.5cm}
    
    \begin{minipage}{\textwidth}
    \centering
    Reconstruction-Based Representations\\
   \includegraphics{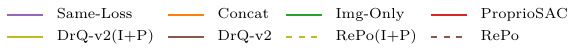}    
   \includegraphics{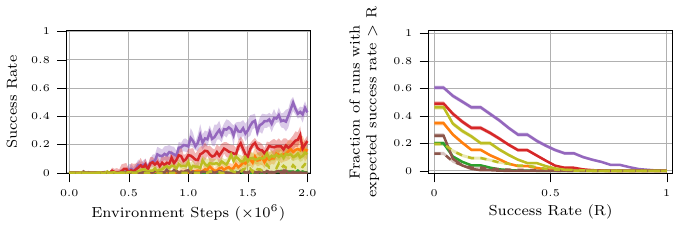}    
    \end{minipage}
    \caption{Aggregated results and performance profiles for the \emph{Manipulation} Suite.
   \emph{\vours} performs best by a significant margin, followed by \emph{\pours}
    No approach that uses solely images, i.e., \emph{Img Only}-ablations, \emph{RePo} and \emph{DrQ-v2}, or uses both modalities but has a fully constrastive objective achieves any notable success.
    }
    \label{fig:mfrl_manip}
\end{figure}

\begin{figure}
    \centering
    \begin{minipage}{\textwidth}
    \centering
    Contrastive Variational Representations\\
     \includegraphics{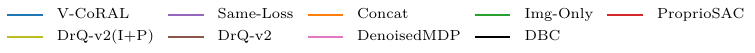} 
    \end{minipage}
    \begin{minipage}{\textwidth}
    \centering
     \includegraphics{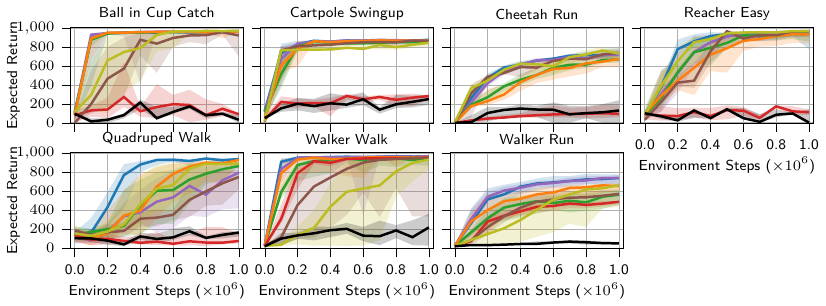}
    \end{minipage}
    
    \vspace{0.4cm}
    
    \begin{minipage}{\textwidth}  
    \centering
    Contrastive Predictive Coding Representations \\
    \includegraphics{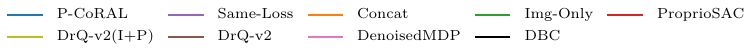}
    \end{minipage}
    \begin{minipage}{\textwidth}
    \centering
     \includegraphics{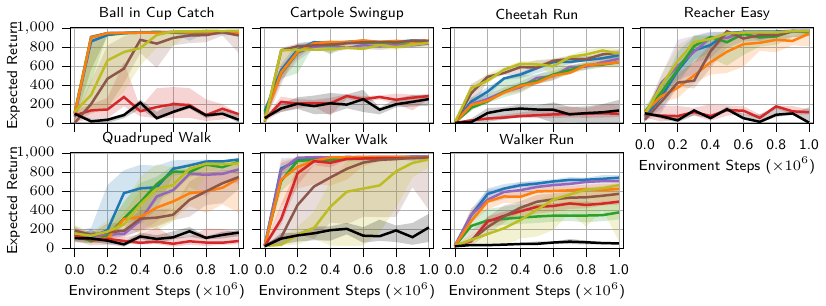}
    \end{minipage}

    \vspace{0.4cm}
    
    \begin{minipage}{\textwidth}
    \centering
    Reconstruction-Based Representation\\
     \includegraphics{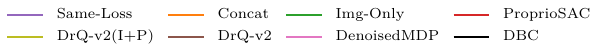}
    \end{minipage}
    \begin{minipage}{\textwidth}
    \centering
     \includegraphics{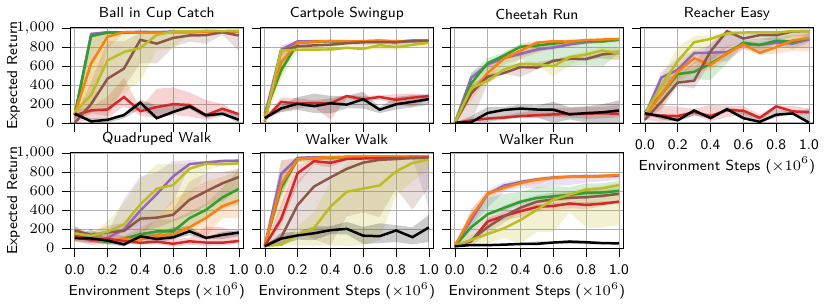}
    \end{minipage}
    
    \caption{Per environment results for model-free agents on the DeepMind Control Suite with \emph{Standard Images}.}
    \label{fig:indi:mfrl_std}
\end{figure}

\begin{figure}
    \centering
    \begin{minipage}{\textwidth}
    \centering
    Contrastive Variational Representations\\
     \includegraphics{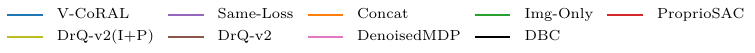}
    \end{minipage}
    \begin{minipage}{\textwidth}
    \centering
     \includegraphics{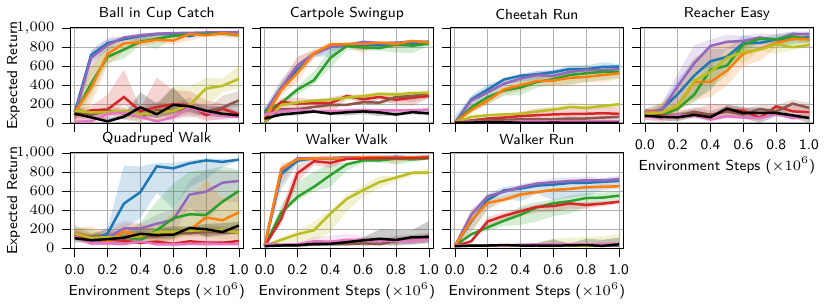}
    \end{minipage}

    \vspace{0.4cm}
    
    \begin{minipage}{\textwidth}
    \centering
    Contrastive Predictive Coding Representations \\
     \includegraphics{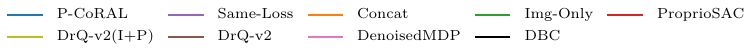}
    \end{minipage}
    \begin{minipage}{\textwidth}
    \centering
     \includegraphics{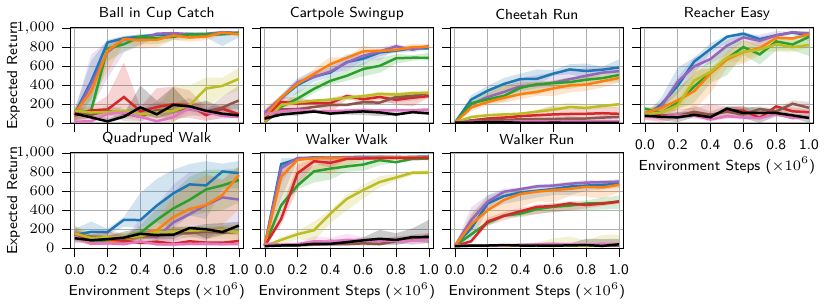}
    \end{minipage}
    
    \vspace{0.4cm}
    
    \begin{minipage}{\textwidth}
    \centering
    Reconstruction-Based Representation\\
     \includegraphics{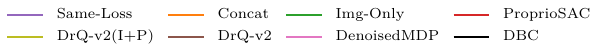}
    \end{minipage}
    \begin{minipage}{\textwidth}
    \centering
     \includegraphics{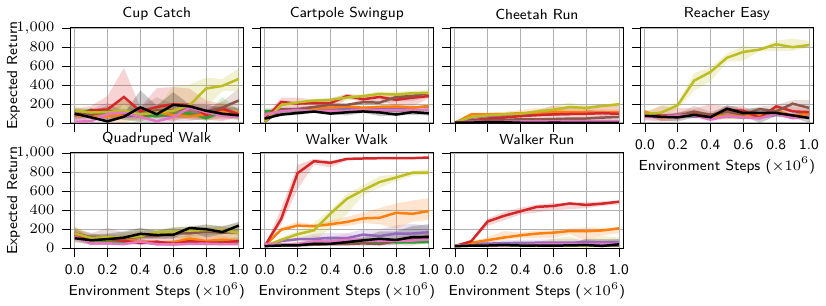}
    \end{minipage}

    \caption{Per environment results for model-free agents on the DeepMind Control Suite with \emph{Video Background}.}
    \label{fig:indi:mfrl_nat}
\end{figure}

\begin{figure}
    \centering
    \begin{minipage}{\textwidth}
    \centering
    Contrastive Variational Representations\\
     \includegraphics{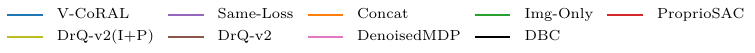}      
    \end{minipage}
    \begin{minipage}{\textwidth}
    \centering
     \includegraphics{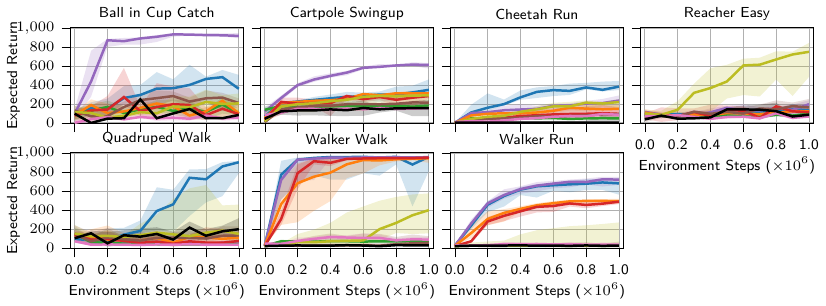}
    \end{minipage}
    
    \vspace{0.4cm}

    \begin{minipage}{\textwidth}
    \centering
    Contrastive Predictive Coding Representations \\
     \includegraphics{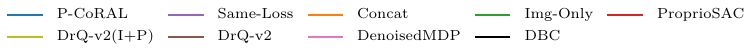}
    \end{minipage}
    \begin{minipage}{\textwidth}
    \centering
     \includegraphics{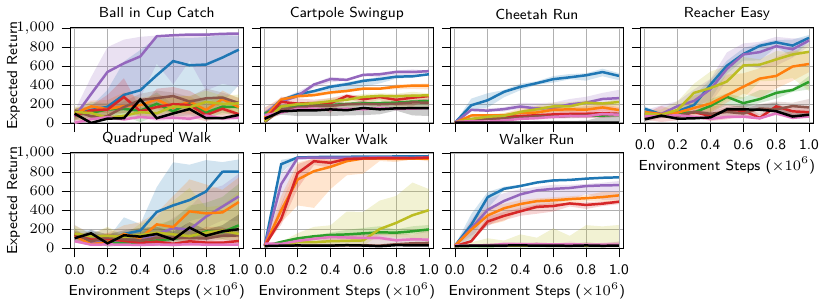}
    \end{minipage}
    
    \vspace{0.4cm}
    
    \begin{minipage}{\textwidth}
    \centering    
    Reconstruction-Based Representation\\
     \includegraphics{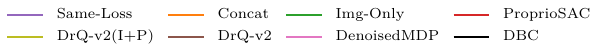}
    \end{minipage}
    \begin{minipage}{\textwidth}
    \centering    
     \includegraphics{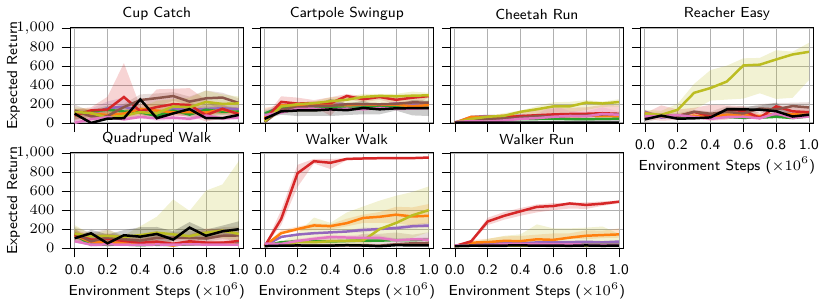}
    \end{minipage}
    
    \caption{Per environment results for model-free agents on the DeepMind Control Suite with \emph{Occlusions}.}
    \label{fig:indi:mfrl_occ}
\end{figure}

\begin{figure}
    \centering
    \begin{minipage}{\textwidth}
    \centering    
    Contrastive Variational Representations\\
     \includegraphics{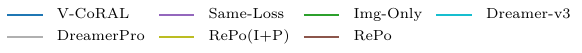}
    \end{minipage}
    \begin{minipage}{\textwidth}
    \centering    
     \includegraphics{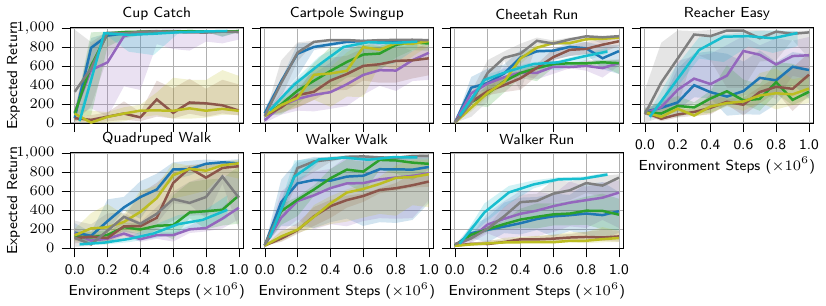}
    \end{minipage}
    
    \vspace{0.4cm}
    
    \begin{minipage}{\textwidth}
    \centering    
    Contrastive Predictive Coding Representations \\
     \includegraphics{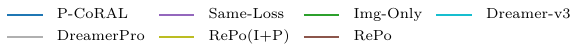}
    \end{minipage}
    \begin{minipage}{\textwidth}
    \centering    
     \includegraphics{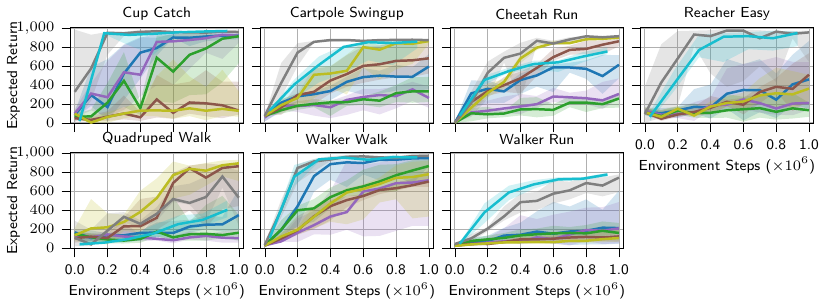}
    \end{minipage}
    
    \vspace{0.4cm}
    
    \begin{minipage}{\textwidth}
    \centering        
    Reconstruction-Based Representation\\
     \includegraphics{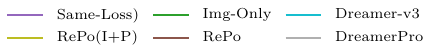}
    \end{minipage}
    \begin{minipage}{\textwidth}
    \centering    
     \includegraphics{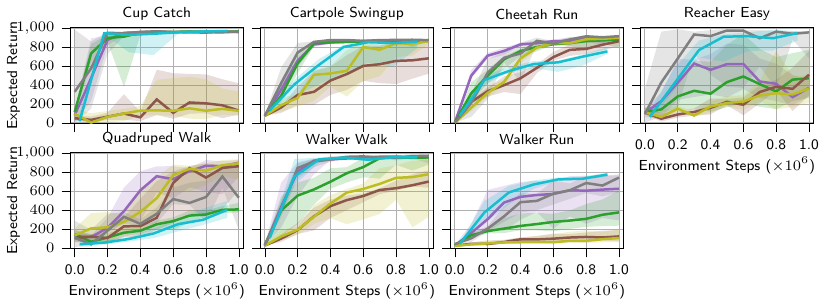}
    \end{minipage}

    \caption{Per environment results for model-based agents on the DeepMind Control Suite with \emph{Standard Images}.}
    \label{fig:indi:mbrl_std}
\end{figure}

\begin{figure}
    \centering
    \begin{minipage}{\textwidth}
    \centering  
    Contrastive Variational Representations\\
     \includegraphics{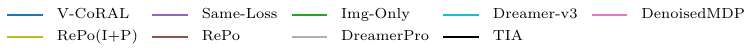}
    \end{minipage}
    \begin{minipage}{\textwidth}
    \centering  
     \includegraphics{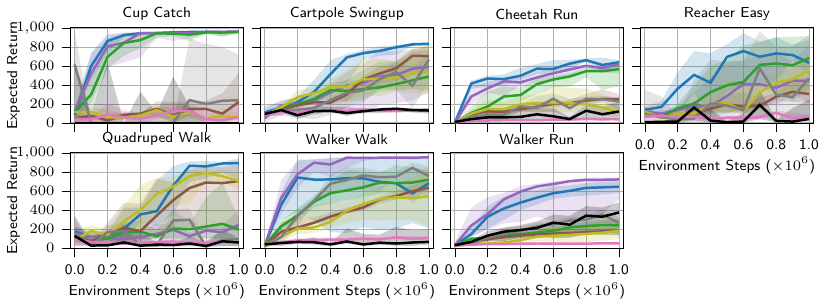}    
    \end{minipage}
    
    \vspace{0.4cm}
    \begin{minipage}{\textwidth}
    \centering  
    Contrastive Predictive Coding Representations \\
     \includegraphics{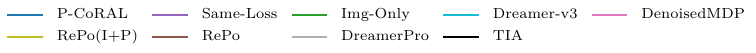}
    \end{minipage}
    \begin{minipage}{\textwidth}
    \centering  
     \includegraphics{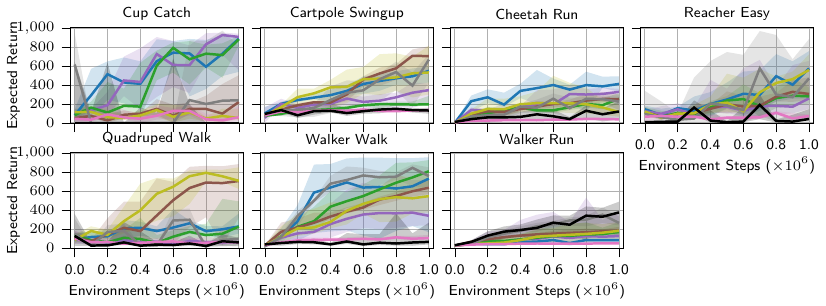}
    \end{minipage}
    
    \vspace{0.4cm}

    \begin{minipage}{\textwidth}
    \centering  
    Reconstruction-Based Representation\\
     \includegraphics{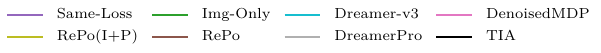}
    \end{minipage}
    \begin{minipage}{\textwidth}
    \centering  
     \includegraphics{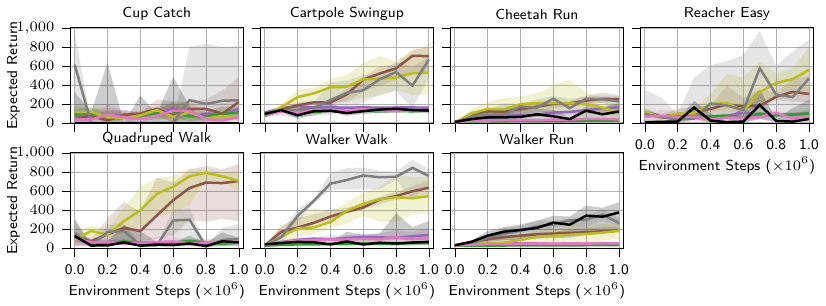}
    \end{minipage}
    \caption{Per environment results for model-based agents on the DeepMind Control Suite with \emph{Video Background}.}
    \label{fig:indi:mbrl_nat}
\end{figure}

\begin{figure}
    \centering
    \begin{minipage}{\textwidth}
    \centering    
    Contrastive Variational Representations\\
     \includegraphics{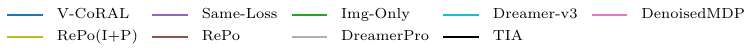}
    \end{minipage}
    \begin{minipage}{\textwidth}
    \centering    
     \includegraphics{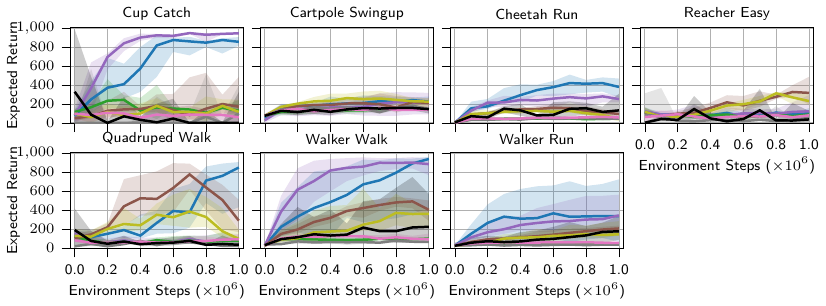}
    \end{minipage}
    
    \vspace{0.4cm}
        
    \begin{minipage}{\textwidth}
    \centering    
    Contrastive Predictive Coding Representations \\
     \includegraphics{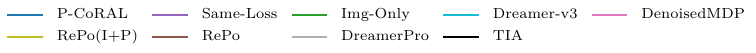}
    \end{minipage}
    \begin{minipage}{\textwidth}
    \centering    
     \includegraphics{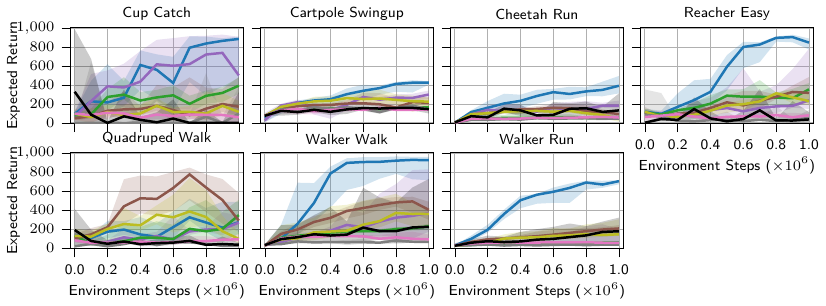}
    \end{minipage}
    
    \vspace{0.4cm}
    
    Reconstruction-Based Representation\\
    \begin{minipage}{\textwidth}
    \centering    
     \includegraphics{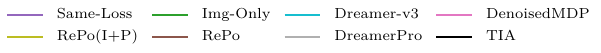}
    \end{minipage}
    \begin{minipage}{\textwidth}
    \centering    
     \includegraphics{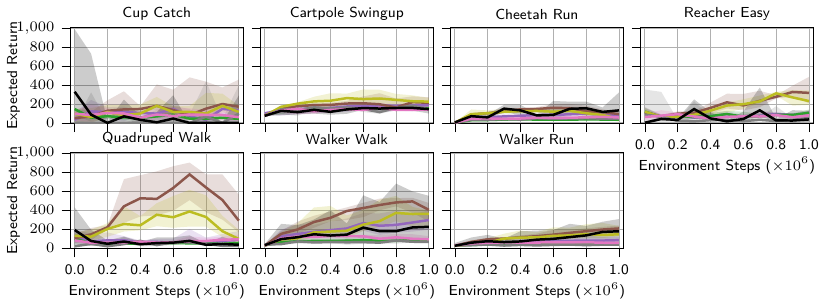}
    \end{minipage}

    \caption{Per environment results for model-based agents on the DeepMind Control Suite with \emph{Occlusions}.}
    \label{fig:indi:mbrl_occ}
\end{figure}

\begin{figure}
    \centering
    \begin{minipage}{\textwidth}
    \centering    
    Contrastive Variational Representations\\
     \includegraphics{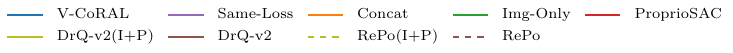}
    \end{minipage}
    \begin{minipage}{\textwidth}
    \centering    
     \includegraphics{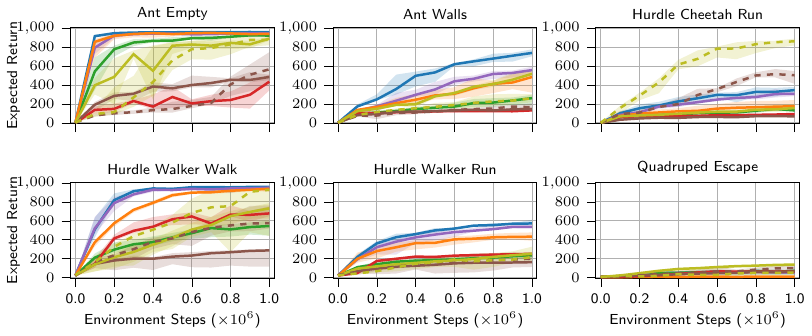}
    \end{minipage}
    
    \vspace{0.4cm}

    \begin{minipage}{\textwidth}
    \centering    
    Contrastive Predictive Coding Representations \\
     \includegraphics{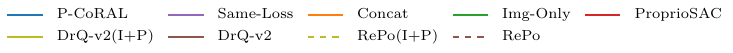}
    \end{minipage}
    \begin{minipage}{\textwidth}
    \centering    
     \includegraphics{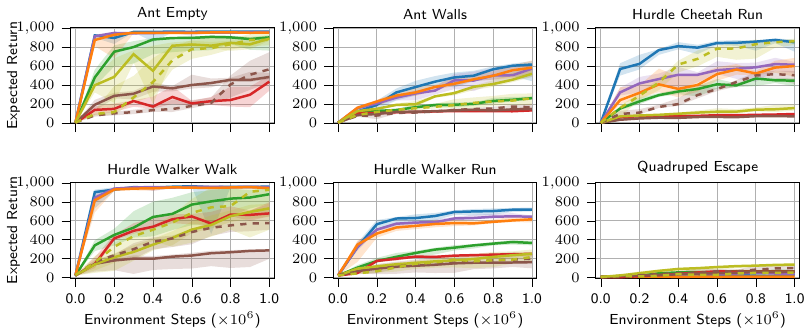}
    \end{minipage}
    
    \vspace{0.4cm}

    \begin{minipage}{\textwidth}
    \centering    
    Reconstruction-Based Representation\\
     \includegraphics{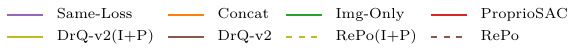}
    \end{minipage}
    \begin{minipage}{\textwidth}
    \centering    
     \includegraphics{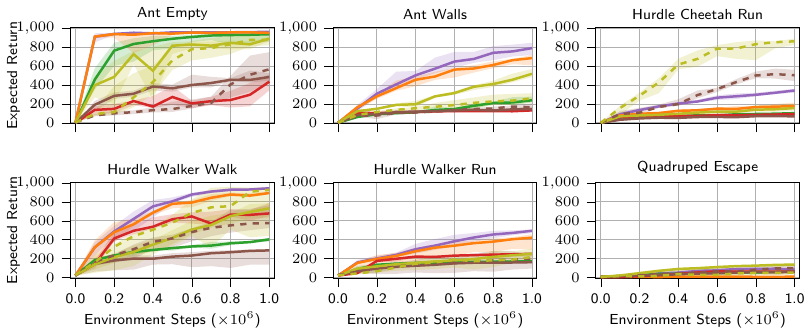}
    \end{minipage}
    
    \caption{Per environment results for the \emph{Locomotion} suite.}
    \label{fig:indi:loco}
\end{figure}

\begin{figure}
    \centering
    \begin{minipage}{\textwidth}
    \centering    
    Contrastive Variational Representations\\
     \includegraphics{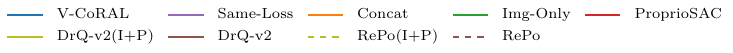}
    \end{minipage}
    \begin{minipage}{\textwidth}
    \centering    
     \includegraphics{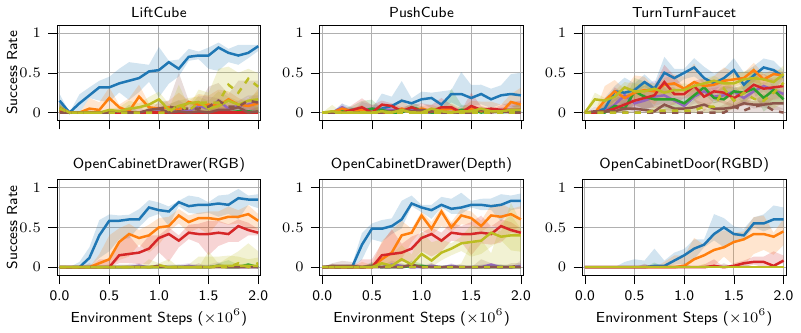}
    \end{minipage}
    
    \vspace{0.4cm}

    \begin{minipage}{\textwidth}
    \centering    
    Contrastive Predictive Coding Representations \\
     \includegraphics{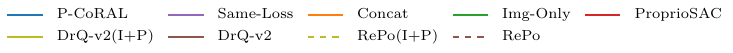}
    \end{minipage}
    \begin{minipage}{\textwidth}
    \centering    
     \includegraphics{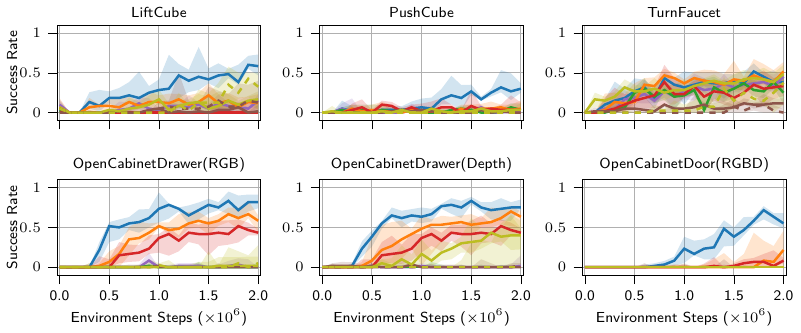}
    \end{minipage}
    
    \vspace{0.4cm}
    
    \begin{minipage}{\textwidth}
    \centering        
    Reconstruction-Based Representation\\
     \includegraphics{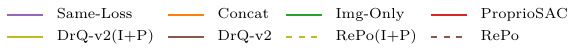}
    \end{minipage}
    \begin{minipage}{\textwidth}
    \centering    
     \includegraphics{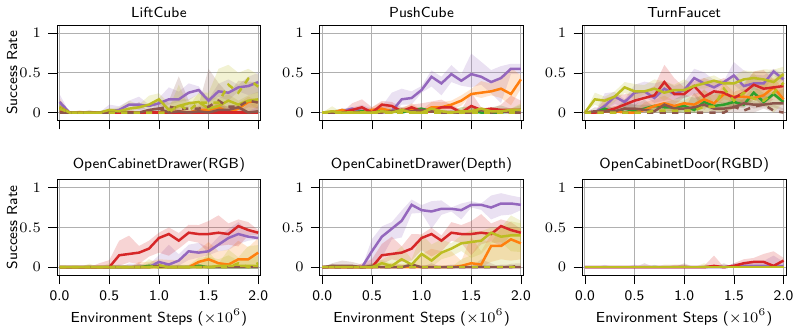}
    \end{minipage}
    
    \caption{Per environment results for the \emph{Manipulation} suite.}
    Reconstruction-Based Representation\\
    \label{fig:indi:manip}
\end{figure}

The following pages list the aggregated results and performance profiles for all tasks, representation-learning approaches, and both model-free and model-based RL. 
We compute inter-quartile means and stratified bootstrapped confidence intervals, as well as the performance profiles according to the recommendations of \citet{agarwal2021deep} using the provided library\footnote{\url{https://github.com/google-research/rliable}}.
For each task in the suites, we ran $5$ seeds per method, i.e., the results for \emph{Standard Images}, \emph{Video Backgrounds}, and \emph{Occlusions} are aggregated over $35$ runs, and those for \emph{Locomotion} over $30$ runs.
 For \emph{OpenCabinetDrawer} we run $20$ seeds per method.
\autoref{fig:mfrl_agg} lists the aggregated results for all model-free agents on the DeepMind Control (DMC) Suite tasks and \autoref{fig:mfrl_pp} lists the corresponding performance profiles. 
\autoref{fig:mbrl_agg} lists the aggregated results for all model-based agents on the DeepMind Control Suite tasks and \autoref{fig:mbrl_pp} lists the corresponding performance profiles. 
\autoref{fig:mfrl_loco} shows aggregated results and performance profiles for the \emph{Locomotion} suite. 
\autoref{fig:mfrl_manip} shows aggregated results and performance profiles for the \emph{Manipulation} suite. 
We also list the per-task results for all task suits:  
\begin{itemize}
    \item \autoref{fig:indi:mfrl_std}: Model-free agents on DMC tasks with \emph{Standard Images} 
    \item \autoref{fig:indi:mfrl_nat}: Model-free agents on DMC tasks with \emph{Video Background}. 
    \item \autoref{fig:indi:mfrl_occ}: Model-free agents on DMC tasks with \emph{Occlusions}.
    \item \autoref{fig:indi:mbrl_std}: Model-based agents on DMC tasks with \emph{Standard Images}. 
    \item \autoref{fig:indi:mbrl_nat}: Model-based agents on DMC tasks with \emph{Video Background}. 
    \item \autoref{fig:indi:mbrl_occ}: Model-based agents on DMC tasks with \emph{Occlusions}. 
    \item \autoref{fig:indi:loco}: Per Environment Results for the \emph{Locomotion} suite.
    \item \autoref{fig:indi:manip}: Per Environment Results for the 
    \emph{Manipulation} suite.
\end{itemize}
\newpage

\end{document}